\documentclass[journal]{IEEEtran}

\usepackage{microtype}

\usepackage{graphicx}
\usepackage{amsmath}
\usepackage{amsfonts}
\usepackage{mathtools}
\usepackage{diagbox}
\usepackage{multirow}
\usepackage{float}
\usepackage{icomma}
\usepackage[range-phrase=--]{siunitx}
\usepackage{algorithm, algpseudocode}
\usepackage{parskip}
\usepackage{tabularx}
\usepackage{booktabs}

\usepackage{xargs}    
\usepackage[pdftex,dvipsnames]{xcolor} 
\usepackage[colorinlistoftodos,prependcaption,textsize=tiny]{todonotes}

\newcommandx{\improvement}[2][1=]{\todo[linecolor=Plum,backgroundcolor=Plum!25,bordercolor=Plum,#1]{#2}}

\usepackage{glossaries}
\newacronym{MLC}{MLC}{multi-label scene classification}
\newacronym{SLC}{SLC}{single-label classification}
\newacronym{RS}{RS}{remote sensing}
\newacronym{CV}{CV}{computer vision}
\newacronym{LP}{LP}{label propagation}
\newacronym{xAI}{xAI}{explainable artificial intelligence}
\newacronym{GAN}{GAN}{generative adversarial network}
\newacronym{SAR}{SAR}{synthetic aperture radar}
\newacronym{CAMs}{CAMs}{class activation maps}
\newacronym{LULC}{LULC}{land use land cover}
\newacronym{RRC}{RRC}{RandomResizeCrop}
\newacronym{RR}{RR}{RandomRotate}

\makeatletter
\newcount\SOUL@minus 
\makeatother

\usepackage{color, soul}

\usepackage{etoolbox}
\usepackage{overpic}

\renewcommand{\arraystretch}{1.4}%
\def\subfigurerowdistance{1em}

\def\subcaptionabove{subcaptionabove}
\def\subcaptionbelow{subcaptionbelow}

\def\subcaptionstyle{subcaptionbelow}

\usepackage{cite}
\usepackage{hyperref}
\usepackage[
    singlelinecheck=false, 
    font=small,
]{caption}
\usepackage[slc=off, subrefformat=parens]{subcaption}

\ifx\subcaptionstyle\subcaptionbelow
\else
\fi

\newcommand{\dynamiccaption}[2]{
    \ifx\subcaptionstyle\subcaptionabove
        \setlength{\belowcaptionskip}{10pt}
        #1
    \fi
    #2
    \ifx\subcaptionstyle\subcaptionbelow
        #1
    \fi
}

\usepackage[capitalise,noabbrev]{cleveref}

\makeatletter
\patchcmd{\@makecaption}
  {\scshape}
  {}
  {}
  {}
\makeatletter
\patchcmd{\@makecaption}
  {\\}
  {.\ }
  {}
  {}
\makeatother

\newcommand*{\linebreaknopagebreak}{%
  \par
  \nobreak
  \vspace{-\parskip}%
  \noindent
  \ignorespaces
}
\begin{document}

\title{\title{A Label Propagation Strategy for CutMix in Multi-Label Remote Sensing Image Classification}}

\author{Tom~Burgert,~\IEEEmembership{~Member,~IEEE,}
        Kai Norman Clasen,~\IEEEmembership{~Member,~IEEE,}
        Jonas~Klotz,~\IEEEmembership{~Member,~IEEE,}
        Tim Siebert,~\IEEEmembership{~Member,~IEEE,}
        and~Begüm~Demir,~\IEEEmembership{Senior~Member,~IEEE}
\thanks{T. Burgert, K. N. Clasen, J. Klotz, T. Siebert, and B. Demir are with the Berlin Institute for the Foundations of Learning and Data (BIFOLD) and with the Faculty of Electrical Engineering and Computer Science, Technische Universit{\"a}t Berlin, 10623 Berlin,
Germany (emails: t.burgert@tu-berlin.de, k.clasen@tu-berlin.de, j.klotz@tu-berlin.de, siebert@tu-berlin.de,  demir@tu-berlin.de). Except for the first and the last author, the order was determined alphabetically.}
}

\markboth{Journal of \LaTeX\ Class Files,~Vol.~14, No.~8, August~2015}%
{Shell \MakeLowercase{\textit{et al.}}: Bare Demo of IEEEtran.cls for IEEE Journals}

\maketitle


\begin{abstract}
The development of supervised deep learning-based methods for multi-label scene classification (MLC) is one of the prominent research directions in remote sensing (RS). However, collecting annotations for large RS image archives is time-consuming and costly. To address this issue, several data augmentation methods have been introduced in RS. Among others, the CutMix data augmentation technique, which combines parts of two existing training images to generate an augmented image, stands out as a particularly effective approach. However, the direct application of CutMix in RS MLC can lead to the erasure or addition of class labels (i.e., label noise) in the augmented (i.e., combined) training image. To address this problem, we introduce a label propagation (LP) strategy that allows the effective application of CutMix in the context of MLC problems in RS without being affected by label noise. To this end, our proposed LP strategy exploits pixel-level class positional information to update the multi-label of the augmented training image. We propose to access such class positional information from reference maps (e.g., thematic products) associated with each training image or from class explanation masks provided by an explanation method if no reference maps are available. Similarly to pairing two training images, our LP strategy carries out a pairing operation on the associated pixel-level class positional information to derive the updated multi-label for the augmented image. Experimental results show the effectiveness of our LP strategy in general (e.g., an improvement of 2–4\% mAP macro compared to standard CutMix) and its robustness in the case of various simulated and real scenarios with noisy class positional information in particular. Code is available at
\href{https://git.tu-berlin.de/rsim/cutmix_lp}{https://git.tu-berlin.de/rsim/cutmix\_lp}.
\end{abstract}

\begin{IEEEkeywords}
CutMix, data augmentation, label propagation strategy, multi-label scene classification, remote sensing.
\end{IEEEkeywords}

\IEEEpeerreviewmaketitle

\section{Introduction}\label{sec:introduction}
\IEEEPARstart{T}{he} growing interest in Earth observation has led to significant technological advancements, contributing to a recent surge in the availability of \gls{RS} image archives. Typically, an RS image scene consists of multiple classes and, thus, can simultaneously be associated with different \gls{LULC} class labels (i.e., multi-labels). Therefore, the development of \gls{MLC} methods that aim to automatically assign multiple labels to each image in an archive attracts growing research interest in the RS community \cite{sumbul_deep_2020}, \cite{li_multi-label_2020}, \cite{yessou_comparative_2020}, \cite{hao_multilabel_2020}. In particular, \gls{MLC} gains popularity due to its capacity to facilitate a quick categorization and understanding of the image scenes. In line with the advances in machine learning, deep neural networks play a crucial role in many state-of-the-art results in \gls{RS} \gls{MLC}. As a result of expansive and time-consuming annotation processes, an important factor in improving the generalization of these networks is the use of data augmentation techniques. Data augmentation refers to the process of creating slightly modified versions of existing training images by applying stochastic transformations to the images, preserving their semantic characteristics. Consequently, deep neural networks trained with augmented images are less prone to overfitting on individual images and, simultaneously, gain a diversified understanding of individual classes as the effective amount of training images is increased. The data augmentation techniques proposed in the literature typically function in a task-agnostic manner. Most of the existing techniques are initially presented in the context of single-label image classification (assigning a single high-level category label to each image), but are also applied in other related image recognition tasks, such as \gls{MLC}, object detection, or semantic segmentation (i.e., pixel-based classification). While a wide range of data augmentation techniques can also be directly applicable to RS MLC, its current use is limited to simple techniques such as image rotations, translations, and flipping \cite{stivaktakis_deep_2019}. 

CutMix \cite{yun_cutmix_2019} stands out as a particularly effective data augmentation technique. It is inspired by the success of the box augmentation strategy CutOut \cite{devries_improved_2017}, \cite{zhong_random_2020}, which removes areas from an image and replaces them with zeros or random noise to enhance model robustness against partially invisible class features (e.g., occlusion, cut-off parts). CutMix takes this concept further by filling informative parts from different images into the uninformative cutout areas, and updating the label accordingly. Consequently, CutMix not only ensures model robustness to partially invisible class features, but also introduces an additional learning signal, enriching the model's understanding of the data. Due to its effectiveness, CutMix has been successfully adopted for \gls{RS} tasks, such as single-label scene classification \cite{lu_rsi-mix_2022}, ship object detection \cite{suo_boxpaste_2022}, and semantic segmentation \cite{wang_ranpaste_2022}.

However, in contrast to simple image transformations that preserve the entire training image (e.g., rotation, flip), applying CutMix in \gls{MLC} problems remains challenging. Unlike object detection and semantic segmentation, there is no direct access to pixel-level class positional information in \gls{MLC} to correctly update the multi-labels. Applying CutMix to multi-label images without a proper label update strategy can result in the introduction of multi-label noise \cite{burgert_effects_2022}. If a class is only present in the cutout area, the application of CutMix can lead to additive noise, as the multi-label would incorrectly indicate the presence of this class. On the other hand, if a new class is contained in the pasted area originating from a different image, the application of CutMix can lead to subtractive noise, as the multi-label would incorrectly indicate the absence of the class. 

To overcome the problem of introducing multi-label noise when directly applying CutMix in \gls{RS} \gls{MLC}, we propose a \gls{LP} strategy that enables the correct update of the multi-label of a CutMix-augmented image. In addition to pairing two images, the \gls{LP} strategy carries out a pairing operation on associated pixel-level class positional information to derive the updated multi-label vector for the augmented image. In detail, we consider two scenarios associated with the availability of pixel-level class positional information, both of which may involve imperfect inputs but are handled robustly by our strategy. In the first scenario, we assume that the multi-labels of the images originate from associated reference maps (i.e., each class label annotated in the reference map is indicated as present in the multi-label). In other words, for each training image, class labels are available not only at image-level but also at pixel-level. These reference maps may vary in reliability depending on how they are obtained. When the associated reference maps are obtained through the inference of a segmentation model (i.e., pixel-based classification model) (e.g., Dynamic World dataset \cite{brown_dynamic_2022}), unreliability in class positional information can originate from inaccurate predictions. When the associated reference maps are obtained from thematic products \cite{paris_novel_2020}, \cite{buttner_corine_2004}, \cite{arino_global_2012}, \cite{bartholome_glc_2002} (e.g., BigEarthNet \cite{sumbul_bigearthnet_2019}), unreliability in class positional information can have various causes, such as (i) different resolutions between thematic product and the considered image; (ii) imprecise co-registration of thematic product and the considered image; (iii) temporal difference between thematic product and the considered image; or (iv) labeling error in the thematic product. In the second scenario, we assume that associated reference maps are not available. Thus, each training image is only annotated with image-level multi-labels. In this case, we propose to leverage post hoc explanation methods to create class explanation masks that attribute each class indicated as present in the multi-label of an image with the respective relevant pixels. The second scenario may also involve imperfect pixel attributions due to imprecise explanations, a condition under which our strategy remains stable. Experimental results on three diverse datasets demonstrate that our LP strategy is effective and robust, even when the external pixel-level class positional information is noisy or not fully reliable.

In summary, our contributions are the following:
 
\begin{itemize}
\item We propose an \gls{LP} strategy that allows the effective application of CutMix to \gls{RS} \gls{MLC} without being affected by multi-label noise.
 \item  To the best of our knowledge, we are the first to explore the benefits of thematic products and explanation methods to provide pixel-level class positional information for correctly updating the multi-labels of a CutMix-augmented image in \gls{RS} \gls{MLC}.
 \item  We conduct extensive experiments considering class positional information with varying levels of reliability. In particular, we show the effectiveness of our \gls{LP} strategy for: (i) reliable class positional information; (ii) a variety of six simulated types of noisy class positional information; (iii) a scenario in which the noisy class positional information is provided by thematic products; and (iv) a scenario in which the noisy class positional information is provided by explanation methods.
\end{itemize}

The remainder of the paper is organized as follows. Section \ref{related} surveys the related data augmentation techniques presented in \gls{CV} and \gls{RS} communities. In Section \ref{methods}, our LP strategy that allows the application of CutMix in the framework of \gls{MLC} problems in \gls{RS} is introduced. Section \ref{dataset_setup} describes the considered multi-label datasets and the experimental setup, while the experimental results are presented in Section \ref{experimental_results}. Finally, in Section \ref{conclusion}, the conclusion of the work is drawn.
\section{Related Work}
\label{related}

According to our knowledge, there are no data augmentation techniques specifically designed and developed for \gls{MLC} problems. Most of the existing data augmentation techniques are considered task-agnostic and thus are directly used in \gls{MLC} problems. Therefore, in this section, we provide a general overview of data augmentation techniques developed for different image recognition tasks in the \gls{CV} and \gls{RS} communities.

In detail, data augmentation techniques play a crucial role for state-of-the-art results in supervised \cite{he_deep_2016}, \cite{tan_efficientnet_2019},  \cite{liu_convnet_2022},
semi-supervised \cite{xie_self-training_2020}, \cite{berthelot_mixmatch_2019}, \cite{sohn_fixmatch_2020}, and self-supervised learning \cite{chen_simple_2020}, \cite{he_momentum_2020}, \cite{caron_emerging_2021} in \gls{CV}. In particular, these techniques demonstrate their effectiveness in \gls{SLC}, but also in related tasks such as object detection \cite{bochkovskiy_yolov4_2020} and semantic segmentation \cite{ghiasi_simple_2021}. Similarly to \gls{CV}, data augmentation techniques significantly improve the performance and generalization capabilities for \gls{RS} tasks. Recently, researchers have explored various approaches to tailor well-established \gls{CV} data augmentation techniques for \gls{RS} tasks, such as \gls{SLC}  \cite{lu_rsi-mix_2022}, \cite{ding_convolutional_2016}, \cite{yan_semi-supervised_2020}, \cite{xiao_progressive_2021}, \cite{zhang_new_2021}, object detection \cite{suo_boxpaste_2022}, \cite{yan_data_2019}, \cite{zheng_using_2019}, \cite{zhang_semi-supervised_2021}, \cite{zhao_improved_2021}, \cite{chen_novel_2022}, or semantic segmentation tasks \cite{wang_ranpaste_2022}, \cite{you_fmwdct_2022}. With the common goal of alleviating overfitting and serving as a form of regularization in small data regimes, data augmentation approaches can be broadly divided into three categories: (i) image modification (Section \ref{image_modifications}); (ii) image generation (Section \ref{image_generation}); or (iii) image pairing (Section \ref{image_pairing}) that also includes CutMix-based approaches. 

\subsection{Image Modification based Data Augmentation}\label{image_modifications}

The first data augmentation techniques developed are based on image modifications \cite{ciregan_multi-column_2012}. These techniques aim to encode invariances in data distortions. Image modifications can be grouped into two variants: (i) geometric transformations and (ii) channel transformations. Geometric transformations involve operations like horizontal flips, random crops, pixel shifts, distortions to orientations, or random erasing (i.e., CutOut) \cite{devries_improved_2017}, \cite{zhong_random_2020}. Conversely, channel transformations include operations like color distortions, contrast and brightness adjustments, or histogram equalization. Both types of transformations are applied successfully to standard \gls{CV} benchmarks such as CIFAR \cite{krizhevsky_learning_2009} and ImageNet \cite{deng_imagenet_2009} (e.g., \cite{krizhevsky_imagenet_2012}, \cite{zagoruyko_wide_2016}, \cite{zoph_learning_2018}).

Many of these image modification techniques, in particular geometric transformations, also belong to the standard training repertoire for various \gls{RS} benchmarks \cite{neumann_-domain_2019}, \cite{bazi_vision_2021}, \cite{wang_empirical_2023}. In addition, Ding et al. \cite{ding_convolutional_2016} find that augmentations of translation, pose synthesis, and speckle noise are beneficial for the task of vehicle classification on \gls{SAR} data. Wei Zhang and Yungang Cao \cite{zhang_new_2021} address the differences between object structures in \gls{RS} \gls{SLC} and \gls{CV} \gls{SLC} (e.g., smaller objects, less connected objects, less central objects) using \gls{CAMs} \cite{oquab_is_2015}. \gls{CAMs} are used to constrain the geometric transformations of cropping and translation by guaranteeing a minimum area of the relevant pixels to be retained. Further, Gaussian noise injection is constrained by weighting it by the inverse of the contribution of pixels to the final prediction.

\subsection{Image Generation based Data Augmentation}\label{image_generation}

In the context of image generation, various data augmentation techniques are proposed to enhance generalization performance by leveraging synthetically generated images. These approaches have in common that they employ a \gls{GAN} for image generation. In particular, \gls{GAN}-based data augmentation techniques show beneficial effects in scarce data regimes \cite{sixt_rendergan_2018}, \cite{wu_dcgan-based_2020} or for few-shot learning tasks \cite{antoniou_data_2018}. Additionally, Geirhos et al. \cite{geirhos_imagenet-trained_2018} improve the robustness of neural networks towards adversarial examples with a \gls{GAN}-based ImageNet stylizing approach that guides the network to prioritize shape over texture.

In \gls{RS}, image generation approaches are proposed mainly for object detection tasks focused on vehicles such as aircrafts, ships, or ground vehicles, in which images are commonly rare and costly to annotate. While Yan et al. create new images by integrating existing 3D models of different ships \cite{yan_data_2019} or aircraft \cite{yan_novel_2019} into real \gls{RS} scenes, Zheng et al. \cite{zheng_using_2019} synthesize realistic ground vehicles by making use of a \gls{GAN}. Furthermore, Xiao et al. \cite{xiao_progressive_2021} extend the integration of 3D models of ships by applying a neural style transfer (NST) network termed Sim2RealNet to eliminate the domain gap between real and simulated images for single-label ship classification.

\subsection{Image Pairing based Data Augmentation}\label{image_pairing}

Data augmentation techniques aligned with the third category involve mixing existing images with appropriate changes to ground reference labels. Mixup \cite{zhang_mixup_2018} creates convex combinations of pairs of images and their corresponding labels, favoring simple linear behavior between training images. Until now, Mixup forms the basis of many state-of-the-art approaches in semi-supervised \cite{berthelot_mixmatch_2019} and label noise robust learning \cite{li_dividemix_2019}. Inspired by the image pairing approach of Mixup, CutMix extends the box augmentation technique of random erasing (i.e., CutOut) by pasting corresponding bounding boxes of other training images into the erased areas. Further extensions of CutMix exist that paste multiple boxes based on the most attentive regions \cite{walawalkar_attentive_2020}. In object detection, the analogue of CutMix is copying object instances (i.e., bounding boxes) and paste-blending them onto diverse backgrounds \cite{dwibedi_cut_2017}. On the other hand, in semantic segmentation, CutMix-inspired approaches commonly termed as copy-paste mechanisms operate at pixel-level \cite{ghiasi_simple_2021}. Moreover, in object detection, Mosaic augmentation further extends CutMix by stitching four resized images together, followed by a random crop that retains parts of every image. 

Similarly, the majority of adoptions of image pairing approaches in \gls{RS} is based on the concepts of the image blending technique Mixup \cite{zhang_mixup_2018}, or the box augmentation technique CutMix \cite{yun_cutmix_2019}. In \gls{SLC}, Lu et al. \cite{lu_rsi-mix_2022} propose RSI-Mix, a variant of CutMix that selects two images of the same category and copy-pastes a selected bounding box from one image to another. Moreover, Yan et al. \cite{yan_semi-supervised_2020} effectively integrate Mixup into a semi-supervised framework for \gls{SLC}. In object detection, Suo et al. \cite{suo_boxpaste_2022} use a variant of CutMix to artificially increase the object density by pasting existing bounding box annotations of ships from different images onto a single target image. Zhao et al. \cite{zhao_improved_2021} introduce a variant of the Mosaic data augmentation technique that prioritizes rare or hard-to-detect class objects and tackles the problem of class imbalances via a joint data augmentation technique that makes use of Mixup, Mosaic, and SMOTE \cite{chawla_smote_2002}. In addition, Zhang et al. \cite{zhang_semi-supervised_2021} propose Object First Mixup, an alteration of Mixup that adjusts the weights of segmented objects and their background in semi-supervised object detection. Finally, in semantic segmentation, You et al. \cite{you_fmwdct_2022} incorporate the copy-paste mechanism (which is the pixel annotation-based equivalent of CutMix) into a semi-supervised framework in which the ground reference pixels of annotated roads are pasted onto unlabeled images. Concurrently, the framework RanPaste introduced by Wang et al. \cite{wang_ranpaste_2022} exploits color jitter and random Gaussian noise for consistency regularization while making use of a CutMix-like pasting mechanism that pastes a bounding box (in contrast to pixels of a segment) of a labeled image onto an unlabeled image. Consequently, a pseudo-label mask for the paired image is inferred from the predictions of a student model.

In contrast to image modification, image generation approaches and Mixup, the application of CutMix in \gls{MLC} tasks can lead to the erasure or addition of class labels in the augmented (i.e., paired) image. To address this problem, we propose an \gls{LP} strategy that enables the correct update of the multi-label of the augmented image for \gls{RS} \gls{MLC}. 
\section{Proposed Strategy}\label{methods}

In this section, we introduce our label propagation (LP) strategy that aims to allow an effective application of CutMix for \gls{MLC} problems in \gls{RS}. Let $\mathcal{D} \coloneqq \{(\boldsymbol{x}_1, \boldsymbol{y}_1), \cdots, (\boldsymbol{x}_N, \boldsymbol{y}_N)\}$ be a multi-label training set, where $N \in \mathbb{N}$ is the number of training images. Each pair $(\boldsymbol{x}_i, \boldsymbol{y}_i)$ in $D$ consists of an image $\boldsymbol{x}_i \in \mathbb{R}^{C \times H \times W}$, where $C$ is the number of image bands with a height $H$ and a width $W$, and its multi-label vector $\boldsymbol{y}_i \in \{0, 1\}^L$, where $L \in \mathbb{N}$ defines the number of classes. Each entry $(\boldsymbol{y}_i)_l$ refers to the presence (i.e., 1) or absence (i.e., 0) of the respective class $l$. We aim to train a deep neural network $h$ such that it generalizes to classify unseen images correctly. We assume that $|D|$ is not sufficiently large and that the data annotation process is costly and time-consuming. To overcome this issue, we adopt the CutMix algorithm due to its proven success in our training procedure to improve the performance of $h$. However, as mentioned before, the labels in \gls{MLC} do not explicitly indicate class positional information. Therefore, applying CutMix to training images annotated with multi-labels can lead to the potential erasure or addition of class labels in the paired image and, thus, may lead to multi-label noise. To address this problem, we propose an \gls{LP} strategy that aims to preserve the correct class information within the multi-label vector of the augmented image to enable an effective application of CutMix in \gls{MLC}. In detail, for accessing pixel-level class positional information that is incorporated by our \gls{LP} strategy into the CutMix process, we consider two scenarios: (i) the usage of reference maps that correspond to the multi-labels, or (ii) the usage of class explanation masks provided by explanation methods if no reference maps are available. In the following, we initially define the algorithm for CutMix in \gls{MLC}. Subsequently, our \gls{LP} strategy for CutMix in \gls{MLC} is introduced for both scenarios. Finally, the section is concluded with a specification of the box creation algorithm as well as the application procedure of CutMix with our \gls{LP} strategy to an \gls{MLC} task.

\subsection{Definition of the CutMix Algorithm}

Let a box (i.e., rectangular region) $R = (r_a, r_b, r_c, r_d)$ be specified through the left lower corner $(r_a, r_b)$ and the right upper corner $(r_c, r_d)$, respectively. Then, the area of the box is denoted as $A_R = (r_c - r_a) (r_d - r_b)$. Furthermore, the binary mask $B_R = (b_{j_{1}j_{2}})_{j_{1}j_{2}} \in \mathbb{R}^{H \times W}$ corresponding to a box $R$ is defined, where  $b_{j_{1}j_{2}}$ is defined for $j_1,j_2=1,...,n$ as follows:
\begin{align}
    \label{eq:binary_mask}
    b_{j_{1}j_{2}} = 
        \begin{cases}
            1, \textrm{ if } r_a \leq j_1 \leq r_c \textrm{ and } r_b \leq j_2 \leq r_d \\
            0, \textrm{ otherwise.}
        \end{cases}
\end{align}

The goal of CutMix is to generate an augmented training image $\boldsymbol{\Tilde{x}}$ with a corresponding multi-label vector $\boldsymbol{\Tilde{y}}$ by combining two existing training images $\boldsymbol{x}_1, \boldsymbol{x}_2$ and their corresponding label vectors $\boldsymbol{y}_1, \boldsymbol{y}_2$. Therefore, the two boxes $R_1$ and $R_2$, and their corresponding binary masks $B_1$ and $B_2$ are generated. To erase $R_1$ from $\boldsymbol{x}_1$, the inverse binary mask $(1 - B_1)$ is combined with $\boldsymbol{x}_1$ through the Hadamard product, as $(1 - B_1) \odot \boldsymbol{x}_1$. In addition, the box $R_2$ is extracted from $\boldsymbol{x}_2$ by applying the Hadamard product between $\boldsymbol{x}_2$ and the binary mask $B_2$, as $B_2 \odot \boldsymbol{x}_2$. To fill the erased area in $\boldsymbol{x}_1$ with the box selected from $\boldsymbol{x}_2$, the two boxes have to be spatially aligned. This alignment is carried out by the image shift operator $T^x_{R_2 R_1}: \mathbb{R}^{C \times H \times W} \to  \mathbb{R}^{C \times H \times W}$. To explain in detail the working principle of the image shift operator for a single image band (i.e., when $C = 1$), let us assume that the unaligned matrices $B_1 \odot \boldsymbol{x}_1$ and $B_2 \odot \boldsymbol{x}_2$ have the following block matrix structures:
\begin{align}
    \label{eq:shift_operator_exmp1}
    B_1 \odot \boldsymbol{x}_1 &= \begin{pmatrix}
        0 & 0 \\
        0 & \boldsymbol{x}_{R_1}
    \end{pmatrix} \\
    B_2 \odot \boldsymbol{x}_2 &= \begin{pmatrix}
        \boldsymbol{x}_{R_2} & 0 \\
        0 & 0
    \end{pmatrix}.
\end{align}
Then, $T_{R_2 R_1}^x$ shifts $\boldsymbol{x}_{R_2}$ to the position of $\boldsymbol{x}_{R_1}$ as: 
\begin{align}
    \label{eq:shift_operator_exmp2}
    T_{R_2 R_1}^x(B_2 \odot \boldsymbol{x}_2) = \begin{pmatrix}
        0 & 0 \\
        0 & \boldsymbol{x}_{R_2}
    \end{pmatrix}.
\end{align}
This shift can be represented by (channel-wise) matrix multiplication with block-identity matrices. To generate the corresponding multi-label vector $\boldsymbol{\Tilde{y}}$, $\boldsymbol{y}_1$ and $\boldsymbol{y}_2$ are weighted by the respective remaining sizes of the paired image parts of $\boldsymbol{x}_1$ and $\boldsymbol{x}_2$. Then, the augmented pair $(\boldsymbol{\Tilde{x}}, \boldsymbol{\Tilde{y}})$ is defined by the two pairing operations:
\begin{align}
    \label{eq:lp_for_slc_cutmix1}
    \boldsymbol{\Tilde{x}} &= (1 - B_1) \odot \boldsymbol{x}_1 + T_{R_2 R_1}^x (B_2 \odot \boldsymbol{x}_2) \\
    \label{eq:lp_for_slc_cutmix2}
    \boldsymbol{\Tilde{y}} &= (1 - A_R) \boldsymbol{y}_1 +  A_R \boldsymbol{y}_2.
\end{align}

It is worth noting that there exist multiple versions of CutMix. In this work, CutMix with unaligned positions of the boxes $R_1$ and $R_2$ is considered to increase the diversity of the augmented images. For further details, we refer the reader to the original CutMix paper \cite{yun_cutmix_2019}.

\subsection{Proposed Label Propagation Strategy for CutMix in Multi-Label Remote Sensing Image Classification}\label{mlc_cutmix}

\begin{figure*}[ht]
    \renewcommand{\arraystretch}{3}
    \centering
    \begin{subcaptiongroup}
        \begin{overpic}[width=.9\linewidth,clip,trim={0 0 0 1.cm}]{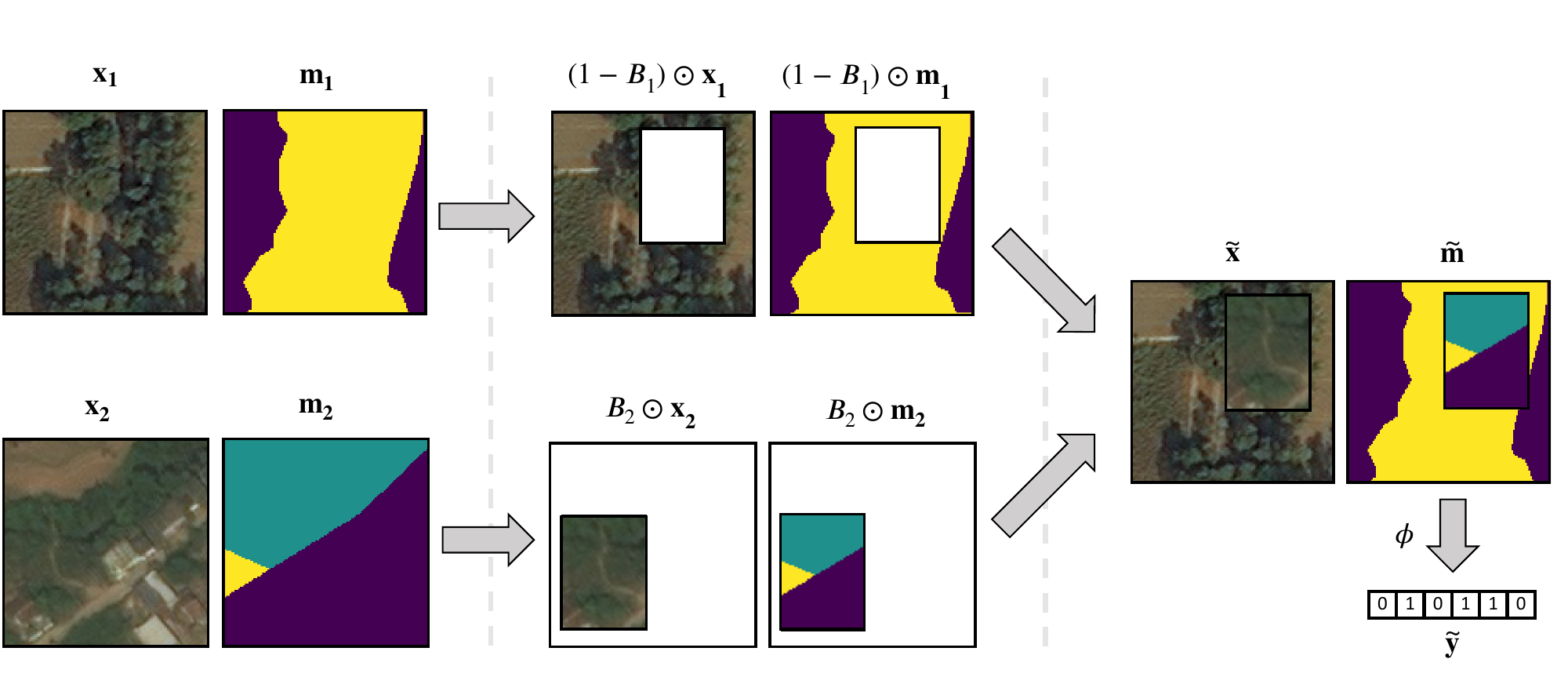}
            \small
            \put(10.6, -1){Step (1)}
            \put(45.5, -1){Step (2)}
            \put(82.5, -1){Step (3)}
        \end{overpic}\\[1em] 
    \end{subcaptiongroup}
    \caption{
    Schematic illustration of the \gls{LP} strategy for CutMix in MLC leveraging reference maps. Step (1): the training images and associated reference maps $(\boldsymbol{x}_1, \boldsymbol{m}_1)$, $(\boldsymbol{x}_2, \boldsymbol{m}_2)$ are selected. Step (2): the binary mask $B_1$ erases a box from $(\boldsymbol{x}_1, \boldsymbol{m}_1)$, and the binary mask $B_2$ extracts a box of the same size from $(\boldsymbol{x}_2, \boldsymbol{m}_2)$, respectively. Step (3): the augmented training image $\boldsymbol{\Tilde{x}}$ and reference map $\boldsymbol{\Tilde{m}}$ are created by leveraging the shift operators to replace the erased area of $(\boldsymbol{x}_1, \boldsymbol{m}_1)$ with the extracted box from $(\boldsymbol{x}_2, \boldsymbol{m}_2)$. The reference map $\boldsymbol{\Tilde{m}}$ is then used to derive the updated multi-label vector $\boldsymbol{\Tilde{y}}$ for $\boldsymbol{\Tilde{x}}$ through the function $\phi$.}
    \label{fig:cutmix_lp_strategy}
\end{figure*}

The proposed \gls{LP} strategy aims to preserve all class labels in the augmented image by leveraging pixel-level class positional information. In detail, we consider two scenarios in which each training image is associated with such class positional information. In the first scenario, the training images are associated with reference maps that directly correspond to the multi-label of the image. In the second scenario, the training images are associated with class explanation masks that are obtained by an explanation method if no reference maps are available. Analogously to the application of the pairing operation for two images in (\ref{eq:lp_for_slc_cutmix1}), our \gls{LP} strategy additionally applies this operation to the pixel-level class positional information of the two images and derives the multi-label vector from the paired information. In the following, the variants of our \gls{LP} strategy for the respective two scenarios to effectively apply CutMix in \gls{MLC} are described.


\textit{Scenario 1: Label Propagation based on Reference Maps} \linebreaknopagebreak

In the first scenario, we assume the existence of associated reference maps. Following the general notation of $[k_1, k_2]_\mathbb{N} \coloneqq \{ j \in \mathbb{N} \;|\; k_1 \leq j \leq k_2\}$ for $k_1, k_2 \in \mathbb{N}$, for each of the pairs $(\boldsymbol{x}_i, \boldsymbol{y}_i)$ in $D$, let $\boldsymbol{m}_i \in [0, L]_\mathbb{N}^{H \times W}$ be a pixel-level reference map associated with the image $\boldsymbol{x}_i$, where each entry in $\boldsymbol{m}_i$ indicates the class label of one of the classes present in $\boldsymbol{y}_i$. In this scenario, we apply a pairing operation analogous to the pairing of $\boldsymbol{x}_1, \boldsymbol{x}_2$ for the two reference maps $\boldsymbol{m}_1, \boldsymbol{m}_2$. Therefore, we define the reference map shift operator $T^m_{R_2 R_1} : [0, L]_\mathbb{N}^{H \times W} \to [0, L]_\mathbb{N}^{H \times W}$. In addition, a read-out function $\phi : [0, L]_\mathbb{N}^{H \times W} \to \{0, 1\}^L$ is defined to extract the class positional information from the updated reference map $\boldsymbol{\Tilde{m}}$ to generate $\boldsymbol{\Tilde{y}}$. In particular, $\phi$ returns the set over all entries (class labels) of the reference map $\boldsymbol{\Tilde{m}}$. Therefore, integrating the \gls{LP} strategy for CutMix  yields the augmented pair $(\boldsymbol{\Tilde{x}}, \boldsymbol{\Tilde{y}})$, where
\begin{align}
    \label{eq:lp_for_mlc_cutmix1}
    \boldsymbol{\Tilde{x}} &= (1 - B_1) \odot \boldsymbol{x}_1 + T_{R_2 R_1}^x (B_2 \odot \boldsymbol{x}_2) \\
    \label{eq:lp_for_mlc_cutmix2}
    \boldsymbol{\Tilde{m}} &= (1 - B_1) \odot \boldsymbol{m}_1 + T_{R_2 R_1}^m (B_2 \odot \boldsymbol{m}_2) \\
    \label{eq:lp_for_mlc_cutmix3}
    \boldsymbol{\Tilde{y}} &= \phi (\boldsymbol{\Tilde{m}}).
\end{align}

An illustration of the proposed \gls{LP} strategy that allows the application of CutMix in multi-label classification problems can be found in Figure \ref{fig:cutmix_lp_strategy}. First, two training images $\boldsymbol{x}_1, \boldsymbol{x}_2$ with their associated reference maps $\boldsymbol{m}_1, \boldsymbol{m}_2$ are selected (Step 1). Then, analogously to CutMix in \gls{SLC}, two boxes $R_1$ and $R_2$ and their corresponding binary masks $B_1$ and $B_2$ are generated to extract boxes from $(\boldsymbol{x}_2, \boldsymbol{m}_2)$ and erase the inverse counterpart boxes from $(\boldsymbol{x}_1, \boldsymbol{m}_1)$ (Step 2). The pairing of the masked images and reference maps is carried out by the shift operators $T^x_{R_2 R_1}$ and $T^m_{R_2 R_1}$. Finally, the multi-label vector $\boldsymbol{\Tilde{y}}$ is read-out by $\phi$ from the updated reference map $\boldsymbol{\Tilde{m}}$ to yield the augmented pair $(\boldsymbol{\Tilde{x}}, \boldsymbol{\Tilde{y}})$ (Step 3).

\textit{Scenario 2: Label Propagation based on Class Explanation Masks} \linebreaknopagebreak

In the second scenario, we assume that there are no associated reference maps available. To address this issue, we propose to use explanation methods to provide the source of pixel-level class positional information that can be leveraged by our \gls{LP} strategy. Let each of the pairs $(\boldsymbol{x}_i, \boldsymbol{y}_i)$ in $D$ be associated with a set of class explanation masks $\boldsymbol{e}_i \in \{0, 1\}^{L \times H \times W}$ generated by an explanation method. An entry of one in $(\boldsymbol{e}_i)_l \in \{0, 1\}^{H \times W}$ corresponds to a pixel (i.e., an activating pixel) in the image that is attributed importance for predicting class $l$. To enable the pairing operation for two sets of class explanation masks $\boldsymbol{e}_1, \boldsymbol{e}_2$, we define the class explanation mask shift operator $T^e_{R_2 R_1}: \{0, 1\}^{L \times H \times W} \to \{0, 1\}^{L \times H \times W}$. In addition, a read-out function $\psi: \{0, 1\}^{L \times H \times W} \to \{0, 1\}^L$ is defined to derive $\Tilde{\boldsymbol{y}}$ from the updated set of class explanation masks $\boldsymbol{\Tilde{e}}$. To address the higher uncertainty between activating and non-activating pixels in border regions of the class explanation masks, this function returns a class $l$ as present in the updated multi-label vector $\boldsymbol{\Tilde{y}}$ if the total number of activating pixels is greater than a minimum pixel threshold $t_\text{map}$. In this setting, individual pixels can be activated by multiple classes, as each class $l$ is represented by an individual binary mask in the set of class explanation masks $\boldsymbol{\Tilde{e}}$. Note that if a class does not belong to the actual multi-label of the corresponding image, the whole explanation mask of this class is set to zero. Thus, it is ensured to only derive reasonable classes in the updated multi-label vector $\boldsymbol{\Tilde{y}}$. Consequently, when leveraging class explanation masks, the augmentation procedure to generate $\Tilde{\boldsymbol{x}}$ is equivalent to (\ref{eq:lp_for_mlc_cutmix1}), while the notation to derive $\Tilde{\boldsymbol{y}}$ from a set of updated class explanation masks $\Tilde{\boldsymbol{e}}$ is similar to (\ref{eq:lp_for_mlc_cutmix2}) and (\ref{eq:lp_for_mlc_cutmix3}), as
\begin{align}
    \label{eq:lp_for_mlc_cutmix_cem1}
    \boldsymbol{\Tilde{e}} &= (1 - B_1) \odot \boldsymbol{e}_1 + T_{R_2 R_1}^e (B_2 \odot \boldsymbol{e}_2) \\
    \label{eq:lp_for_mlc_cutmix_cem2}
    \boldsymbol{\Tilde{y}} &= \psi (\boldsymbol{\Tilde{e}}).
\end{align}
It is worth noting that the class explanation masks can be provided by any explanation method. Although some explanation methods directly output class explanations in the form of binary masks, other methods only provide class explanation heatmaps, where each pixel is associated with a probability of contributing to the prediction of the respective class, e.g., DeepLift \cite{shrikumar_learning_2017} or GradCAM \cite{selvaraju_grad-cam_2017}. In this case, we propose to convert the heatmaps into a binary mask using a threshold $t_\text{cam}$. In practice, this setup requires a two-stage training procedure: 1) pretraining the model without CutMix to obtain class explanation masks through an explanation method, and 2) training from scratch with CutMix and our \gls{LP} strategy using the generated masks. The parameterization and experimental details of this procedure are provided in \Cref{experimental_results}.

\textit{Generation of Boxes in CutMix} \linebreaknopagebreak

It is necessary to generate different boxes with various side lengths and areas throughout the CutMix process. Thus, to meet the requirement of controlling the box sizes, the implementation of CutMix needs to incorporate an input of a minimal area $A_{\min} > 0$ and a maximal area $A_{\max} \geq A_{\min} > 0$ of the box, denoted as the box size range $A_{min}-A_{max}$. Thus, for a box $R = (r_a, r_b, r_c, r_d)$ the points are required to fulfill the inequality:
\begin{align}
    \label{eq:rect_ineq}
    A_{\min} \leq (r_c - r_a) (r_d - r_b) \leq A_{\max}.
\end{align}
We propose an algorithm to create the points that satisfy (\ref{eq:rect_ineq}). It operates batch-wise by creating coordinates $(r_a, r_b, r_c, r_d)$ for $N_R$ boxes, which is the number of boxes needed for the batch-wise application of CutMix. The first step is to sample a list of box coordinates $\mathcal{B}$ uniformly from $[0, H]_\mathbb{N} \times [0, W]_{\mathbb{N}}$. Thus, it is iterated until $N_R$ points satisfy (\ref{eq:rect_ineq}). If there are fewer than $N_R$ points, every pair of points that violates the condition is resampled. The pseudo-code of the algorithm is presented in Algorithm \ref{alg:rectangles}.

\begin{algorithm}[ht!]
\caption{Creation of Boxes}
\label{alg:rectangles}
\begin{algorithmic}[1]
\Require $A_{\min} > 0$; $A_{\max} \geq A_{\min} > 0$; $N_R, H, W \in \mathbb{N}$.
\Ensure List of $N_R$ valid random boxes.
\State Initialize empty list of boxes $\mathcal{B}$
\While{$|\mathcal{B}| < N_R$}
\State Sample $\bar{r}_a$, $\bar{r}_c$ uniformly from $[0, H]_\mathbb{N}$
\State Sample $\bar{r}_b$, $\bar{r}_d$ uniformly from $[0, W]_{\mathbb{N}}$
\State $r_a = \min(\bar{r}_a, \bar{r}_c)$; $r_c = \max(\bar{r}_a, \bar{r}_c)$
\State $r_b = \min(\bar{r}_b, \bar{r}_d)$; $r_d = \max(\bar{r}_b, \bar{r}_d)$
\State Calculate area $A = (r_c-r_a)(r_d-r_b)$
\If{$A_{\min} \leq A \leq A_{\max}$}
\State Append ($r_a$, $r_b$, $r_c$, $r_d$) to $\mathcal{B}$
\EndIf 
\EndWhile
\State \Return $\mathcal{B}$
\end{algorithmic}
\end{algorithm}

\textit{Multi-Label Scene Classification with CutMix and the proposed Label Propagation Strategy} \linebreaknopagebreak

In the training phase of a deep neural network $h$, CutMix is applied to the selected images with a probability $p$. In each training step, after loading a batch from $\mathcal{D}$, random images $\boldsymbol{x}_1$ and $\boldsymbol{x}_2$ with their respective multi-label vectors $\boldsymbol{y}_1$ and $\boldsymbol{y}_2$ and reference maps $\boldsymbol{m}_1$ and $\boldsymbol{m}_2$ (or class explanation masks $\boldsymbol{e}_1$ and $\boldsymbol{e}_2$) are selected for the augmentation process guided by our \gls{LP} strategy. The process yields a replacement of $\boldsymbol{x}_1$ and $\boldsymbol{y}_1$ by probability $p$ with the augmented image $\boldsymbol{\Tilde{x}}$ and its updated multi-label vector $\boldsymbol{\Tilde{y}}$ that is derived by our \gls{LP} strategy based on a pairing operation of the reference maps. It is worth noting that the one-to-one replacement ensures that the batch size is persistent. After training the deep neural network $h$ with CutMix-augmented images together with our proposed \gls{LP} strategy, the predicted multi-labels are assigned to images in the considered archive. 
\section{Dataset Description and Experimental Design} \label{dataset_setup}

\begin{figure*}[t!]
    \centering
    \begin{subfigure}{.25\linewidth}
        \dynamiccaption{\caption{}\label{subfigure:example-deepglobe}}{\includegraphics[width=\linewidth]{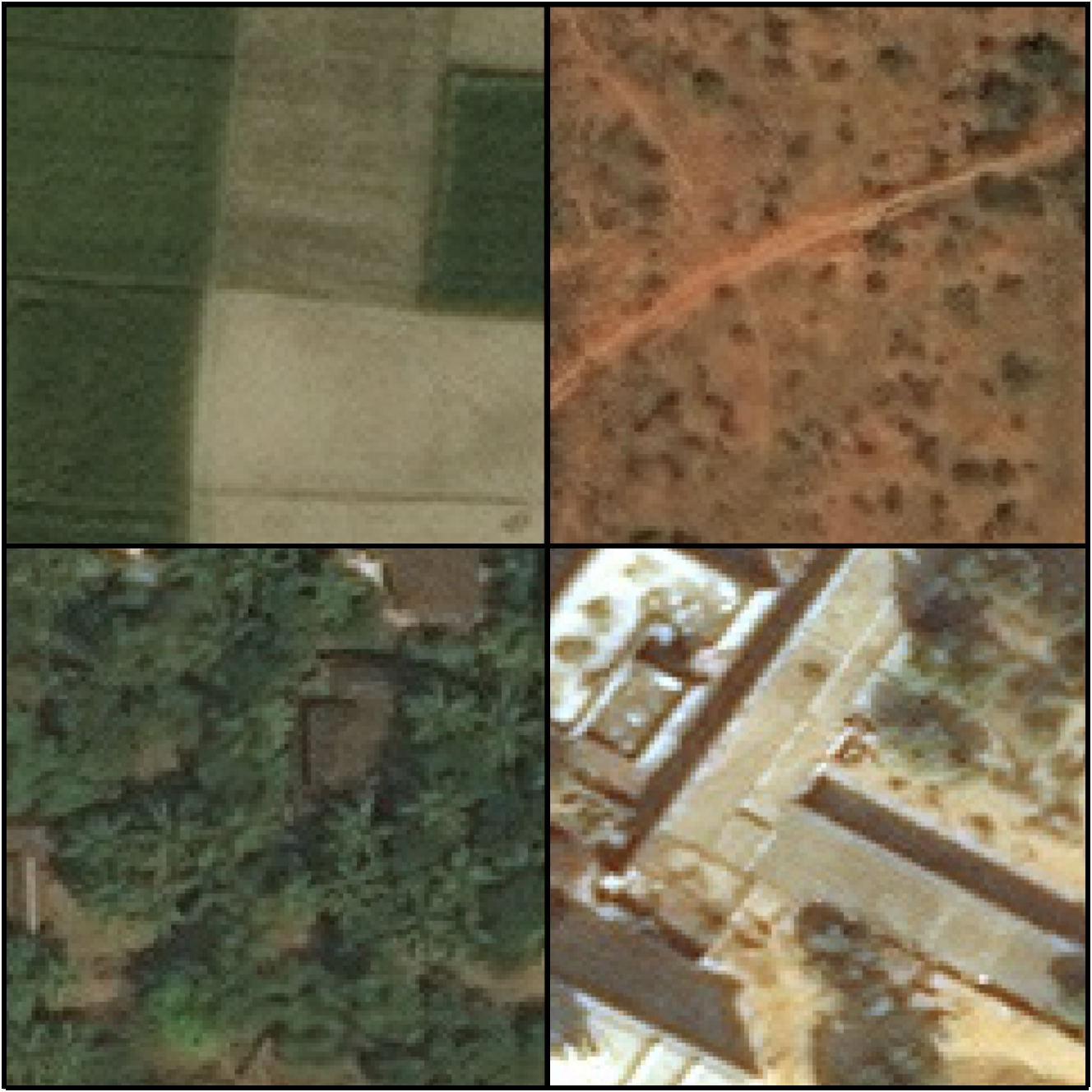}}
    \end{subfigure}\hfil
    \begin{subfigure}{.25\linewidth}
        \dynamiccaption{\caption{}\label{subfigure:example-ben19}}{\includegraphics[width=\linewidth]{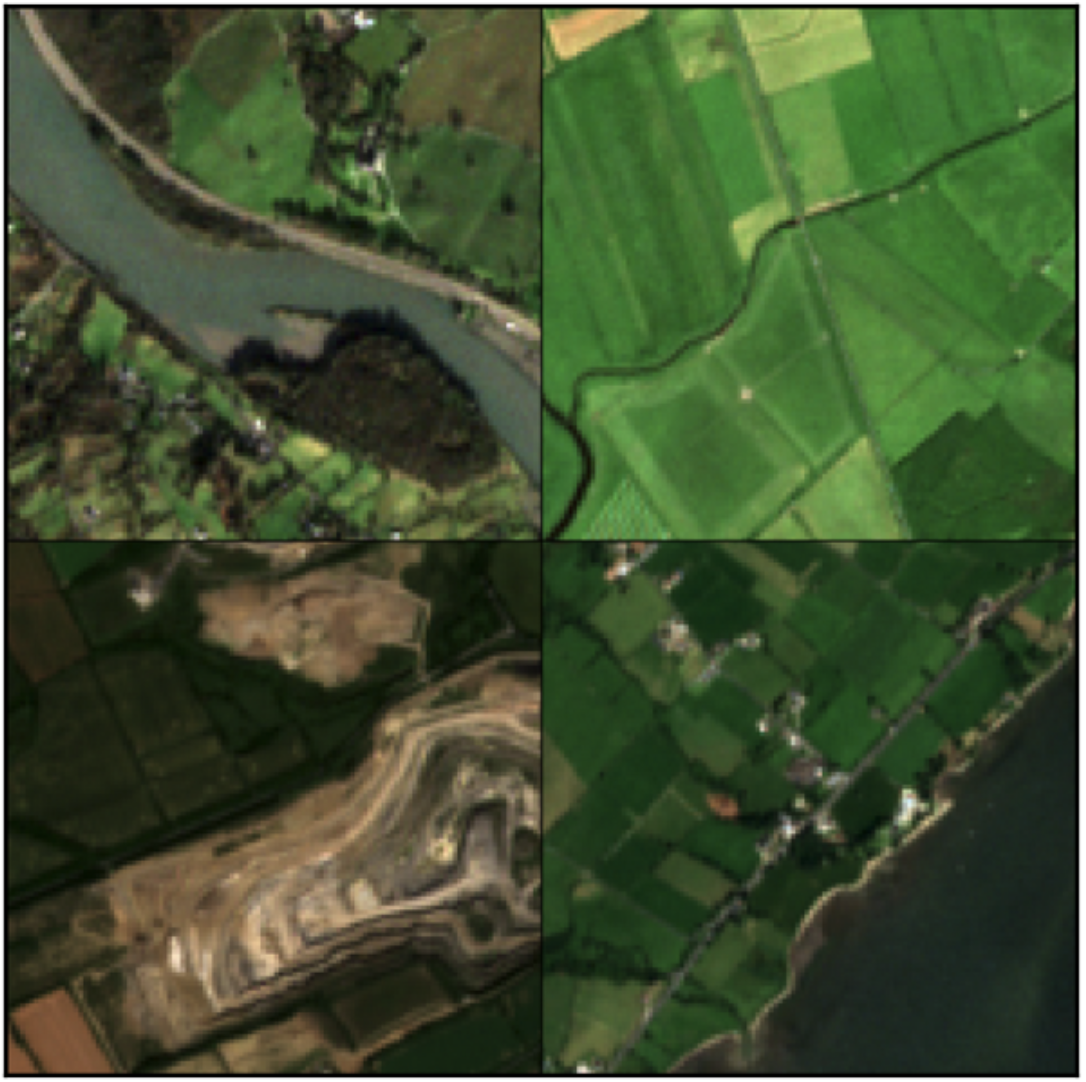}}
    \end{subfigure}\hfil
    \begin{subfigure}{.25\linewidth}
        \dynamiccaption{\caption{}\label{subfigure:example-treesat}}{\includegraphics[width=\linewidth]{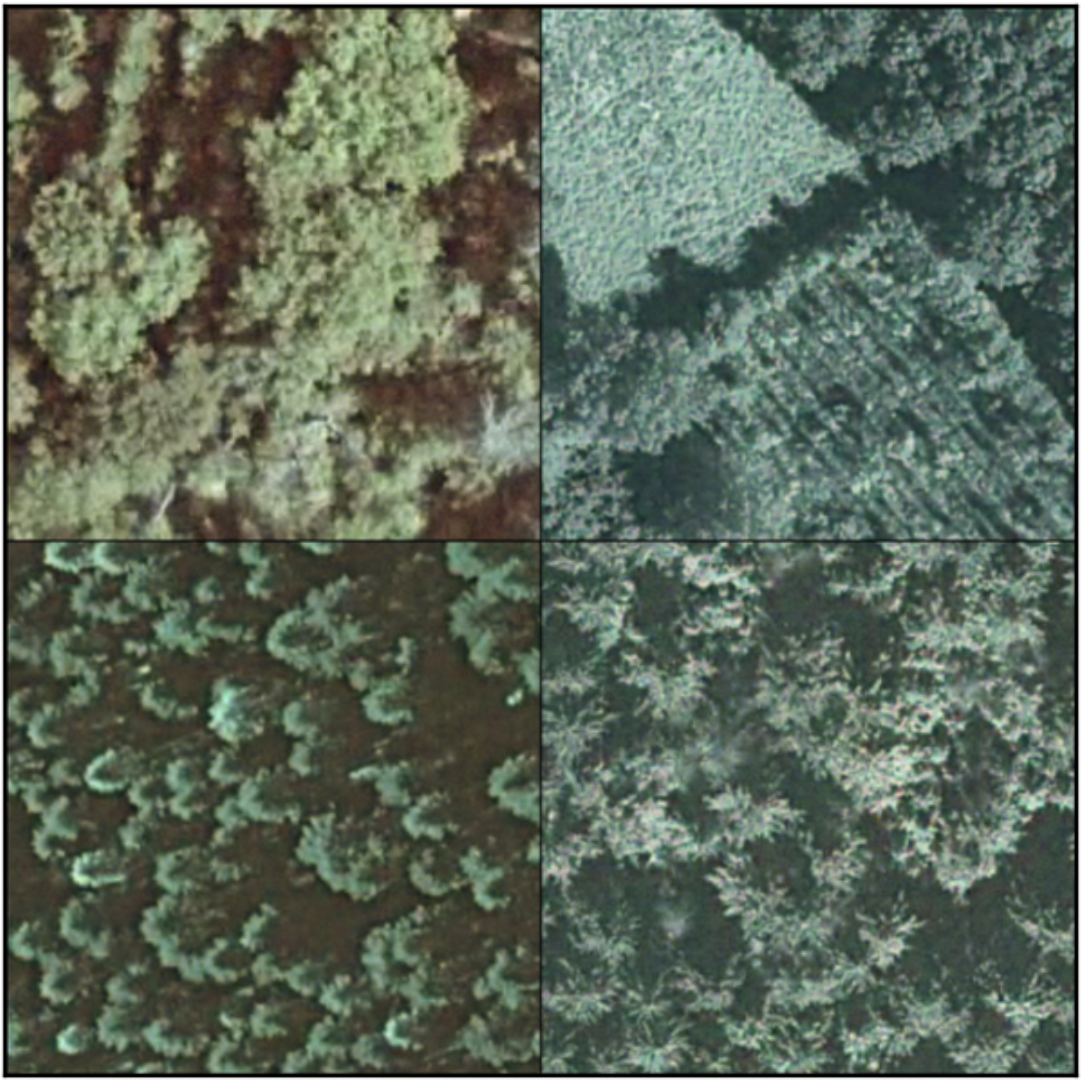}}
    \end{subfigure}
    \caption{Example images of the datasets: \subref{subfigure:example-deepglobe} DeepGlobe-ML, \subref{subfigure:example-ben19}  BigEarthNet-S2, and \subref{subfigure:example-treesat} TreeSatAI.}
    \label{fig:example_images}
\end{figure*}

\begin{table*}[t!]
\caption{Characteristics of the multi-label scene classification datasets used in our experiments.}
\label{table:ds_characteristics}
    \centering
    \renewcommand*{\arraystretch}{1.2}
    \sisetup{group-separator={,}, input-ignore={,}}
     \begin{tabularx}{\linewidth}{p{3cm} S[table-format=6] S[table-format=2] S[table-format=1.2] S[table-format=2] >{\centering\arraybackslash}X r}
     \toprule
    \multirow{2}{*}{Dataset} & {\multirow{2}{*}{$|\mathcal{D}|$}} & {\multirow{2}{*}{\hspace{1.2em}$L$}\hspace{1.2em}} & {Avg. $L$}       &  {\multirow{2}{*}{\hspace{1.2em}$C$\hspace{1.2em}}} & {\multirow{2}{*}{Image Sizes (Spatial Resolution)}} & {\multirow{2}{*}{Pixel-Level Reference Maps}} \\
     & & & {per Image} &  &  &  \\ 
     \midrule
    DeepGlobe-ML & 18,185 & 6 & 1.71 & 3 & 120$\times$120 (0.5m) & Manually Annotated \\ 
     BigEarthNet-S2 \cite{sumbul_bigearthnet_2019} &  250,249 & 19 & 2.95 & 10 & 120$\times$120 (10m), 60$\times$60 (20m) & Thematic Product \\ 
     TreeSatAI \cite{ahlswede_treesatai_2022} & 40,332 & 15 & 1.59 & 4 & 304$\times$304 (0.2m) & Not available \\
     \bottomrule
     \\
    \end{tabularx}
\end{table*}

In the experiments, we have used three RS multi-label datasets: 1) DeepGlobe-ML, which is a multi-label dataset that we constructed from the DeepGlobe Land Cover Classification Challenge dataset \cite{demir_deepglobe_2018}, 2) BigEarthNet-S2 \cite{sumbul_bigearthnet_2019}, and 3) TreeSatAI \cite{ahlswede_treesatai_2022}. For each dataset, example images are shown in \cref{fig:example_images}. As it is common in \gls{MLC}, the datasets used in the experiments exhibit imbalanced class distributions. A comparison of the main characteristics of the considered datasets is provided in \cref{table:ds_characteristics}, while the datasets are briefly described in the following. For the details on these datasets (e.g., statistics on class frequencies), we refer the reader to the original dataset publications \cite{demir_deepglobe_2018}, \cite{sumbul_bigearthnet_2019}, \cite{ahlswede_treesatai_2022}. 

\subsection{Datasets}

\subsubsection{DeepGlobe-ML}
The original DeepGlobe Land Cover Classification Challenge dataset \cite{demir_deepglobe_2018} is collected from the DigitalGlobe +Vivid Images dataset that contains 1949 RGB tiles of size $2448\times2448$ pixels with a spatial resolution of \SI{0.5}{\metre} acquired over Thailand, Indonesia, and India. Each tile is associated with a manually annotated ground reference map. The classes comprise urban, agriculture, rangeland, forest, water, barren, and unknown. To ensure one \gls{MLC} dataset associated with reliable reference maps in this paper, we constructed the DeepGlobe-ML from the original pixel-level classification dataset. To this end, all tiles were divided into a grid of $120\times120$-pixel patches. The respective multi-labels were then derived from the label information of the associated $120\times120$-pixel reference maps. We discarded all patches containing the class unknown. To construct an appropriate \gls{MLC} dataset, we included all patches containing more than one present class. In addition, we retained only \SI{20}{\percent} of the patches with a single present class, resulting in an average of $1.71$ present classes per patch (cf. $1.28$ without discarding patches). In total, DeepGlobe-ML includes $30,443$ patches and is split into a training set (\SI{60}{\percent}), a validation set (\SI{20}{\percent}), and a test set (\SI{20}{\percent}). \\

\subsubsection{BigEarthNet-S2} 
BigEarthNet-S2 \cite{sumbul_bigearthnet_2019} is a multi-label dataset that includes $590,326$ Sentinel-2 multispectral images acquired over ten countries in Europe. The \gls{LULC} class annotations are obtained from reference maps that originate from the publicly available thematic product CORINE Land Cover Map inventory of 2018 \cite{buttner_corine_2004}. Following the \gls{LULC} class nomenclature proposed in \cite{sumbul_bigearthnet-mm_2021}, each image is annotated with a subset of $19$ \gls{LULC} classes, including different types of forests, water, or complex urban or agricultural classes. The dataset contains an average of $2.95$ present classes per image. In the experiments, a filtered subset that excludes patches with seasonal snow, clouds, and cloud shadows is used. The selected subset is divided into a training set (\SI{50}{\percent}), a validation set (\SI{25}{\percent}), and a test set (\SI{25}{\percent}). Furthermore, only $10$ bands (out of the originally $13$ bands) with a spatial resolution of \SI{10}{\metre} ($120\times120$-pixel) and \SI{20}{\metre} ($60\times60$-pixel) are considered. 

\subsubsection{TreeSatAI}
The TreeSatAI dataset \cite{ahlswede_treesatai_2022} is a multi-label tree species classification benchmark dataset that consists of $50,381$ image-triplets of a high-resolution aerial image, a Sentinel-1 SAR, and a Sentinel-2 multispectral image acquired over the German federal state of Lower Saxony. In our experiments, we only use the aerial images. Each of the $304\times304$-pixel aerial images includes the RGB and a near-infrared band with a spatial resolution of \SI{0.5}{\metre} and is associated with multi-labels representing different species of trees from 15 genus classes. The labels were derived from forest administration data from the federal state of Lower Saxony that was annotated in the field or by photo interpretation. For the experiments, we chose a threshold of \SI{5}{\percent} coverage of the pixels for a class to be annotated as present in the image. In addition, we update the original train-test split by also considering a validation set, resulting in a training set of \SI{80}{\percent}, a validation set of \SI{10}{\percent}, and a test set of \SI{10}{\percent} of the images. On average, there are $1.59$ different tree species present in each image. In particular, there exist no other pixel-level reference maps associated with this dataset. The application of our \gls{LP} strategy is only possible when using explanation methods (i.e., generating class explanation masks).

\subsection{Experimental Setup}
For all our experiments, we use a randomly initialized ResNet18 from the \texttt{pytorch-image-models} library \cite{wightman_pytorch_2019}. We choose a batch size of 300 and the cosine annealing learning rate scheduling with a start rate set to 5e-4 and warm-up iterations based on the number of steps. Each model is trained for 120 epochs. For reproducibility, the global random state is fixed to seeds in the range of 42 to 46 implemented by the \texttt{seed-everything} functionality in \texttt{PyTorch lightning} \cite{falcon_pytorch_2019}. We conduct every experiment with these five seeds and report the average of the test set performance in mean average precision macro (denoted as mAP macro) of the best model determined by the validation set. All experiments are carried out on an A100 SXM4 80GB GPU.
\section{Experimental Results} \label{experimental_results}

The experimental results are organized as follows. In Section \ref{subsec:clean_segmaps}, we assess the effectiveness of our LP strategy for CutMix in \gls{RS} \gls{MLC} when reliable pixel-level reference maps are available. Subsequently, in Section \ref{subsec:noisy_segmaps}, we conduct an extensive ablation study on the effect of noisy reference maps on our \gls{LP} strategy. To this end, we define and simulate noise types that can be related to different types of labeling errors. To further show the benefits in real use-cases beyond simulated noisy reference maps, in Section \ref{subsec:thematic_products_segmaps}, we study the effects of our \gls{LP} strategy in a scenario in which the class positional information is extracted from noisy pixel-level reference maps derived from a thematic product. In Section \ref{subsec:xai_segmaps}, we show the advantage of our \gls{LP} strategy in a scenario in which we do not have access to reference maps and instead leverage the class positional information provided by an explanation method. Finally, in Section \ref{subsec:other_das} we compare the effectiveness of CutMix with our \gls{LP} strategy to other data augmentation techniques.

\subsection{Label Propagation based on Reliable Reference Maps}\label{subsec:clean_segmaps}

To evaluate the potential of leveraging pixel-level reference maps to correctly update the multi-labels during the application of CutMix in \gls{RS} \gls{MLC}, we initially study the effects of our proposed \gls{LP} strategy when reliable reference maps are available. To this end, we report the results on the DeepGlobe-ML dataset. We compare three different setups: i) CutMix without our proposed \gls{LP} strategy, denoted as CutMix~w/o~LP; ii) CutMix with our proposed \gls{LP} strategy based on reference maps, denoted as CutMix~w~$\text{LP}_\text{map}$; and iii) CutMix with our proposed \gls{LP} strategy based on class explanation masks, denoted as CutMix~w~$\text{LP}_\text{xAI}$ (see \cref{subsec:xai_segmaps}). As these techniques may favor different box sizes, we conduct experiments for the value of the box size range $A_{min}-A_{max}$ as \numrange{0.1}{0.3}, \numrange{0.1}{0.5}, \numrange{0.1}{0.7}, \numrange{0.3}{0.5}, and \numrange{0.3}{0.7}. All techniques are applied with a probability $p$ of 0.5. As a baseline, we include the performance of the model trained without any augmentation techniques. The results are shown in \cref{table:results_deepglobe_clean}. From the table, it can be seen that for each box size CutMix with our proposed \gls{LP} strategy improves the application of CutMix w/o \gls{LP}. In particular, it can be observed that the larger the selected box sizes become, the more beneficial it is to apply our \gls{LP} strategy. While for the box size range of \numrange{0.3}{0.7}, CutMix~w~$\text{LP}_\text{map}$ results in the best overall performance of all methods with \SI{82.30}{\percent} in mAP macro, the box size range also yields one of the largest improvements to CutMix w/o LP. This improvement can be attributed to the highest probability of deleting class labels from one of the paired images.

\begin{table}[h!]
\caption{Results in mAP macro (\si{\percent}) for DeepGlobe-ML with associated reliable reference maps.}
\label{table:results_deepglobe_clean}
    \centering
     \begin{tabularx}{\linewidth}{X S S S S S}
     \toprule
     \multicolumn{6}{c}{\textbf{CutMix Augmentation}} \\
     \midrule
     & \multicolumn{5}{c}{Box Size Range} \\
     \cmidrule{2-6}
     Method & \numrange{0.1}{0.3} & \numrange{0.1}{0.5} & \numrange{0.1}{0.7} & \numrange{0.3}{0.5} & \numrange{0.3}{0.7} \\ 
     \midrule
     CutMix w/o LP & 78.93 & 79.17 & 79.76 & 80.25 & 80.49 \\ 
     {CutMix~w~$\text{LP}_\text{map}$} & \textbf{80.33} & \textbf{80.74} & \textbf{80.72} & \textbf{82.21} & \textbf{82.30} \\
     {CutMix~w~$\text{LP}_\text{xAI}$} & \textbf{79.54} & \textbf{79.70} & \textbf{80.24} & \textbf{81.96} & \textbf{82.01} \\
     \midrule
     \multicolumn{6}{c}{\textbf{No Augmentation}} \\
     \midrule
     Baseline & 78.40 & & & & \\
     \bottomrule
     \\
    \end{tabularx}
\end{table}

\subsection{Label Propagation based on Simulated Noisy Reference Maps}\label{subsec:noisy_segmaps}

\begin{figure*}[t!]
    \centering
    \def\segnoisefrac{.24}
    \begin{subfigure}{\segnoisefrac\linewidth}
        \dynamiccaption{\caption{}\label{subfigure:seg_noise_original}}{\includegraphics[width=\linewidth]{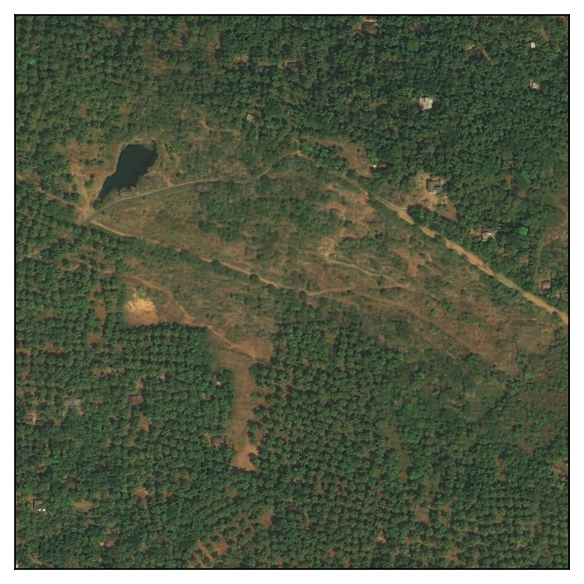}}
    \end{subfigure}
    \begin{subfigure}{\segnoisefrac\linewidth}
        \dynamiccaption{\caption{}\label{subfigure:seg_noise_original_seg}}{\includegraphics[width=\linewidth]{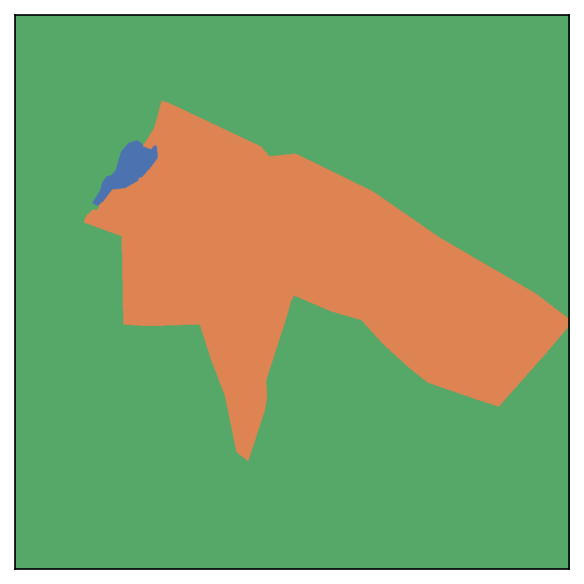}}
    \end{subfigure}
    \begin{subfigure}{\segnoisefrac\linewidth}
        \dynamiccaption{\caption{}\label{subfigure:seg_noise_shift}}{\includegraphics[width=\linewidth]{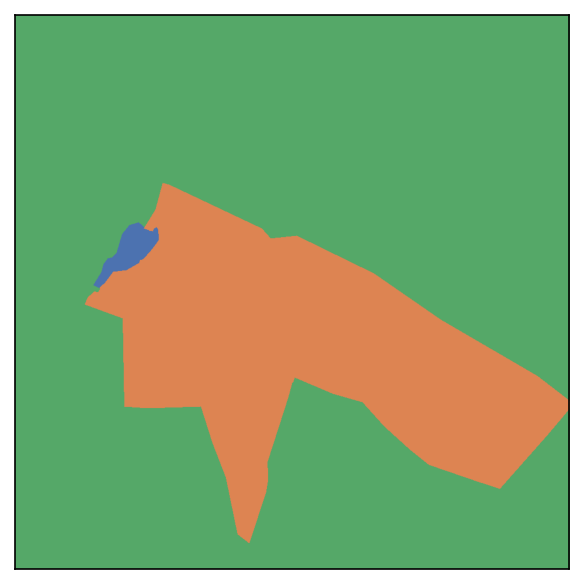}}
    \end{subfigure}
    \begin{subfigure}{\segnoisefrac\linewidth}
        \dynamiccaption{\caption{}\label{subfigure:seg_noise_dilation}}{\includegraphics[width=\linewidth]{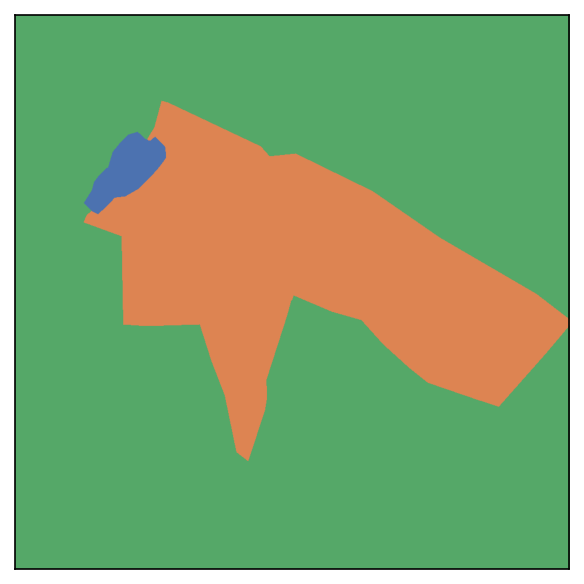}}
    \end{subfigure}
    \\\vspace{\subfigurerowdistance}
    \begin{subfigure}{\segnoisefrac\linewidth}
        \dynamiccaption{\caption{}\label{subfigure:seg_noise_rectify}}{\includegraphics[width=\linewidth]{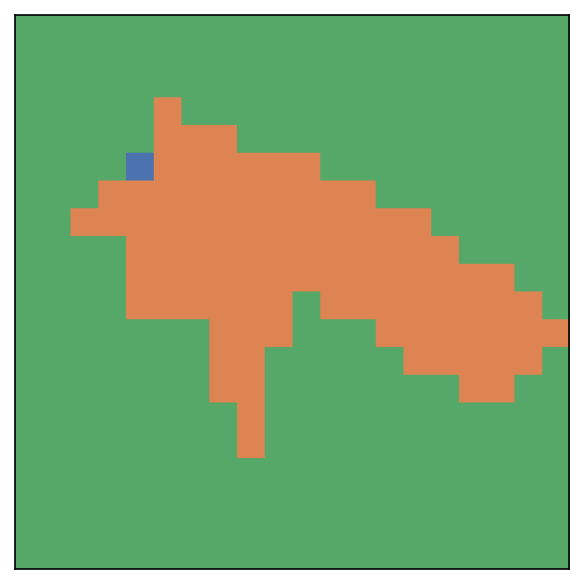}}
    \end{subfigure}
    \begin{subfigure}{\segnoisefrac\linewidth}
        \dynamiccaption{\caption{}\label{subfigure:seg_noise_border_deform}}{\includegraphics[width=\linewidth]{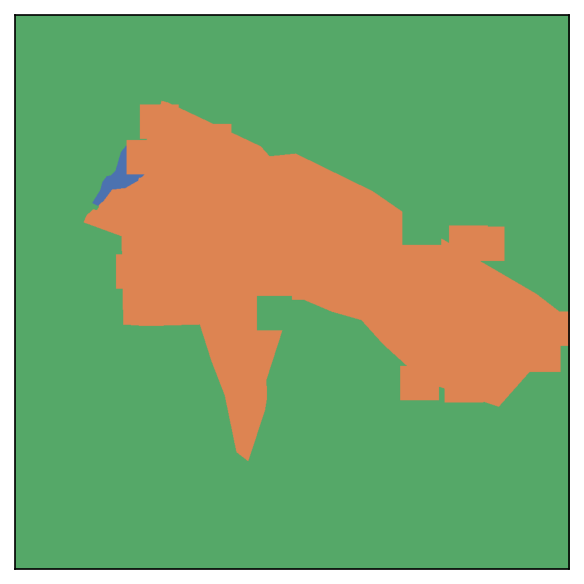}}
    \end{subfigure}
    \begin{subfigure}{\segnoisefrac\linewidth}
        \dynamiccaption{\caption{}\label{subfigure:seg_noise_segment_swap}}{\includegraphics[width=\linewidth]{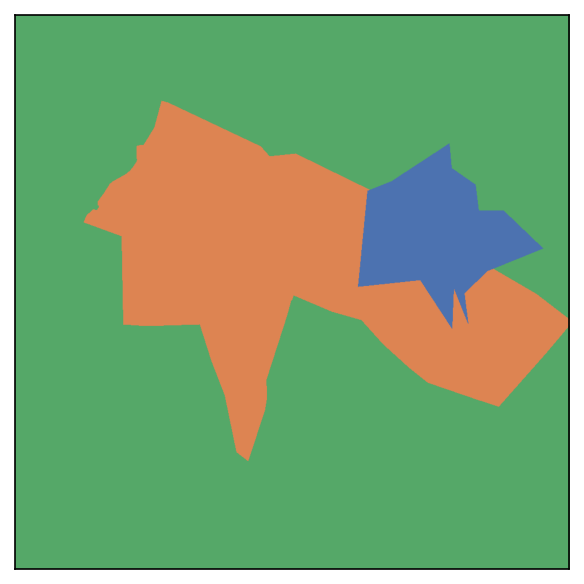}}
    \end{subfigure}
    \begin{subfigure}{\segnoisefrac\linewidth}
        \dynamiccaption{\caption{}\label{subfigure:seg_noise_class_swap}}{\includegraphics[width=\linewidth]{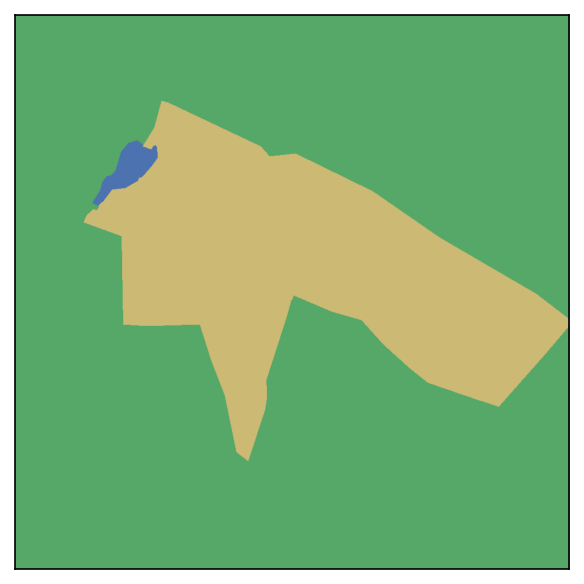}}
    \end{subfigure}
    \caption{Examples of simulated noise in pixel-level reference maps: \subref{subfigure:seg_noise_original}~original image, \subref{subfigure:seg_noise_original_seg}~original reference map, \subref{subfigure:seg_noise_shift}~Mask Shift, \subref{subfigure:seg_noise_dilation}~Dilation/Erosion, \subref{subfigure:seg_noise_rectify}~Rectify Borders, \subref{subfigure:seg_noise_border_deform}~Border Deformation, \subref{subfigure:seg_noise_segment_swap}~Segment Swap, and \subref{subfigure:seg_noise_class_swap}~Class Swap.}
    \label{fig:seg_noise_types}
\end{figure*}

Only in rare cases do reliable pixel-level reference maps exist.
In many scenarios, multi-labels are associated with not fully reliable
reference maps, and sometimes even the estimation of class positional information via explanation methods is necessary. To assess the robustness of our \gls{LP} strategy, we perform an ablation study with six types of simulated noise in reference maps: (i) Mask Shift, (ii) Dilation/Erosion, (iii) Rectify Borders, (iv) Border Deformation, (v) Segment Swap, and (vi) Class Swap. These settings test the generalizability of the \gls{LP} strategy to arbitrary sources of noisy class positional information. In the following, we provide intuitive examples for each noise type, which can often be linked to multiple scenarios. The illustrations are shown in \cref{fig:seg_noise_types}, with the example image in \cref{subfigure:seg_noise_original} and the corresponding reference map in \cref{subfigure:seg_noise_original_seg}.

\textit{Mask Shift}. This noise type reflects an imperfect alignment between the reference map and the image (see \cref{subfigure:seg_noise_shift}), most prominently due to imprecise co-registration of image and label source when derived from a thematic product. To simulate this type of noise, a fraction $f$ of reference maps is shifted in a random direction by a uniformly sampled number of pixels up to $\alpha_\text{pixel\_shift}$.

\textit{Dilation/Erosion}. This noise type affects the borders of a class segment that are proportionally enlarged or reduced (see \cref{subfigure:seg_noise_dilation}), for example, due to a temporal mismatch between image and label source when derived from a thematic product, e.g., a dried lake or burned forest. To simulate this type of noise, a fraction $f$ of reference maps has a randomly selected class segment that is either dilated or eroded for $\alpha_\text{iterations}$ iterations.

\textit{Rectify Borders}. This noise type represents coarse reference maps (see \cref{subfigure:seg_noise_rectify}), e.g., when the label source from a thematic product has a lower resolution than the image. To simulate this type of noise, a fraction $f$ of reference maps is downsampled by $\alpha_\text{magnitude}$ and then resized back to the original resolution.

\textit{Border Deformation}. This noise type is associated with
inaccurate borders between class segments of the reference map (see \cref{subfigure:seg_noise_border_deform}), for example, due to human annotation errors, inaccurate predictions from a segmentation model, or temporal changes such as new urban structures. To simulate this noise type, $\alpha_\text{max\_def}$ are placed on each reference map for a fraction $f$ of the training set. If a box overlaps multiple segments, the whole box is assigned to a randomly chosen bordering class. Otherwise, the segment remains unchanged.

\textit{Segment Swap}. This noise type reflects an incorrect positional assignment of a class (see \cref{subfigure:seg_noise_segment_swap}), which can occur when explanation methods incorrectly localize a class. To simulate this type of noise, a fraction $f$ of reference maps has one class segment removed and randomly placed in an aleatory polygon shape at a different location in the same reference map.

\textit{Class Swap}. This noise type represents mislabeling of class segments (see \cref{subfigure:seg_noise_class_swap}), e.g., human annotation errors in the thematic product for an ambiguous class or incorrect predictions of a segmentation model. To simulate this type of noise, a fraction $f$ of all segments in the training reference maps are reassigned to a uniformly sampled new class. Since this induces multi-label noise, the associated image-level labels are updated accordingly for comparison of CutMix with and without \gls{LP}.

\begin{figure*}[t!]
    \centering
    \def\simnoisefrac{.325}
    \begin{subfigure}{\simnoisefrac\linewidth}
        \dynamiccaption{\caption{}\label{subfigure:sim_noise_shift_result}}{\includegraphics[width=\linewidth]{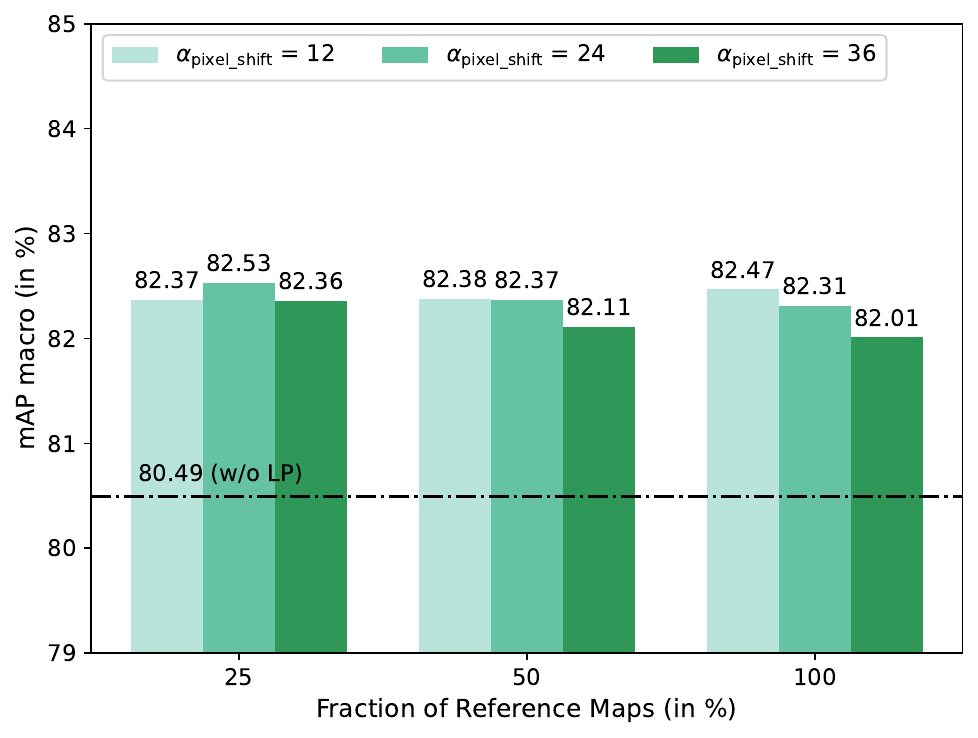}}
    \end{subfigure}
    \begin{subfigure}{\simnoisefrac\linewidth}
        \dynamiccaption{\caption{}\label{subfigure:sim_noise_dilation_result}}{\includegraphics[width=\linewidth]{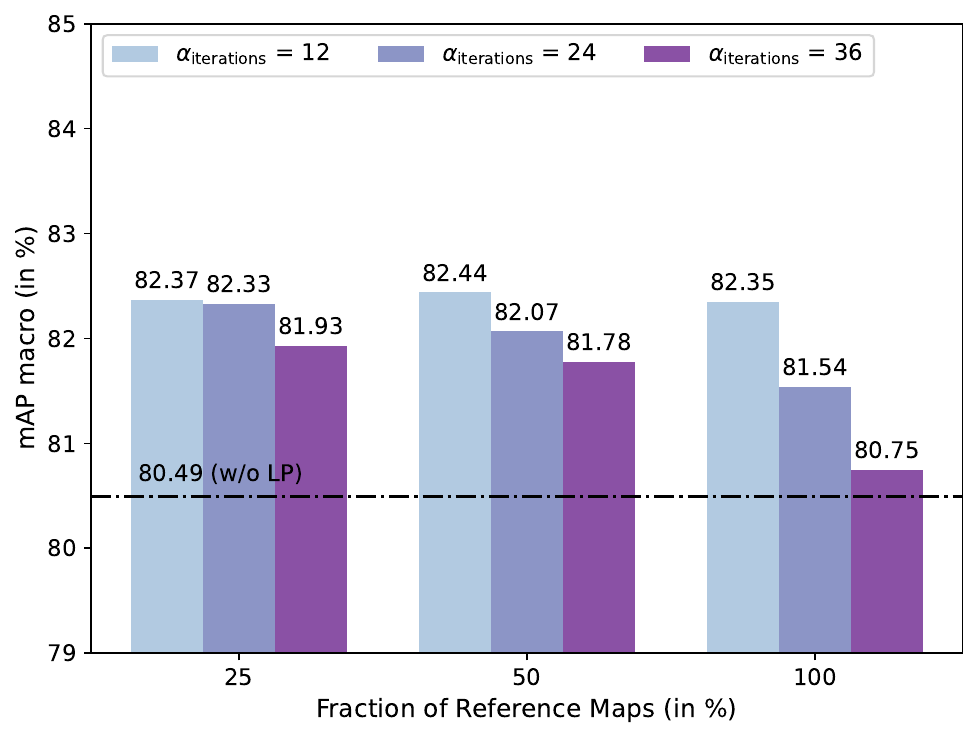}}
    \end{subfigure}
    \begin{subfigure}{\simnoisefrac\linewidth}
        \dynamiccaption{\caption{}\label{subfigure:sim_noise_rectify_result}}{\includegraphics[width=\linewidth]{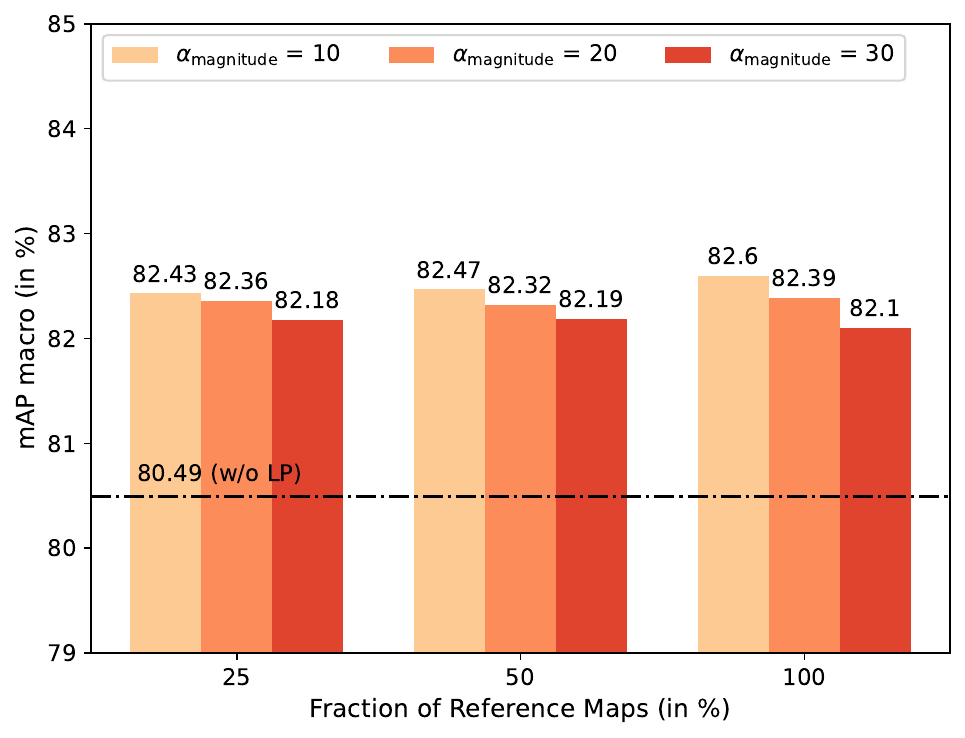}}
    \end{subfigure}
    \\\vspace{\subfigurerowdistance}
    \begin{subfigure}{\simnoisefrac\linewidth}
        \dynamiccaption{\caption{}\label{subfigure:sim_noise_border_deform_result}}{\includegraphics[width=\linewidth]{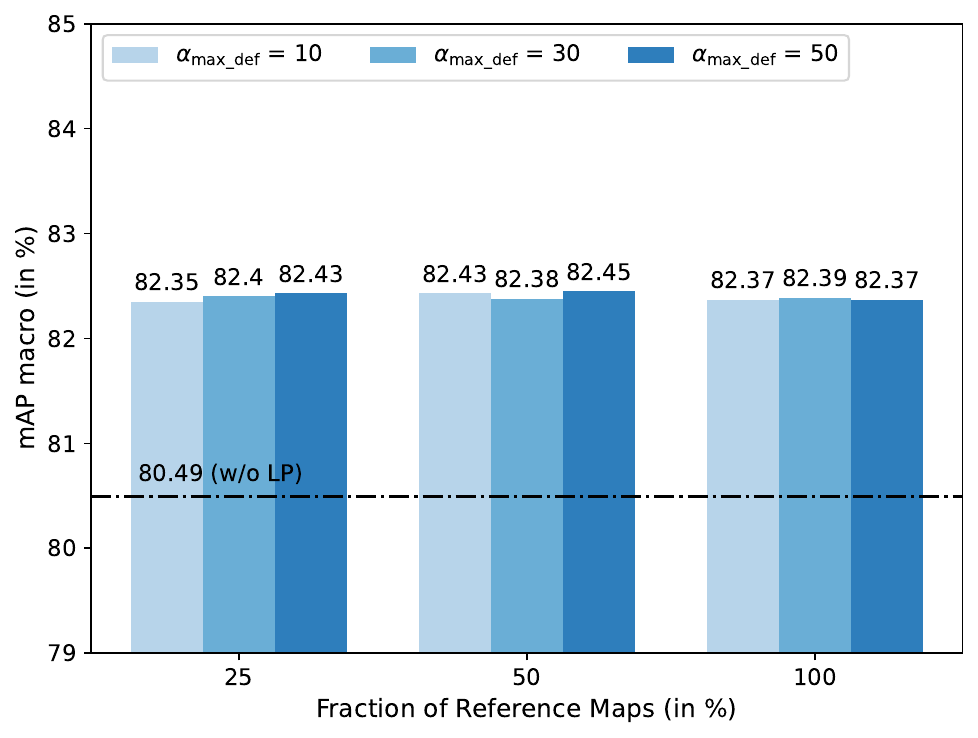}}
    \end{subfigure}
    \begin{subfigure}{\simnoisefrac\linewidth}
        \dynamiccaption{\caption{}\label{subfigure:sim_noise_segment_swap_result}}{\includegraphics[width=\linewidth]{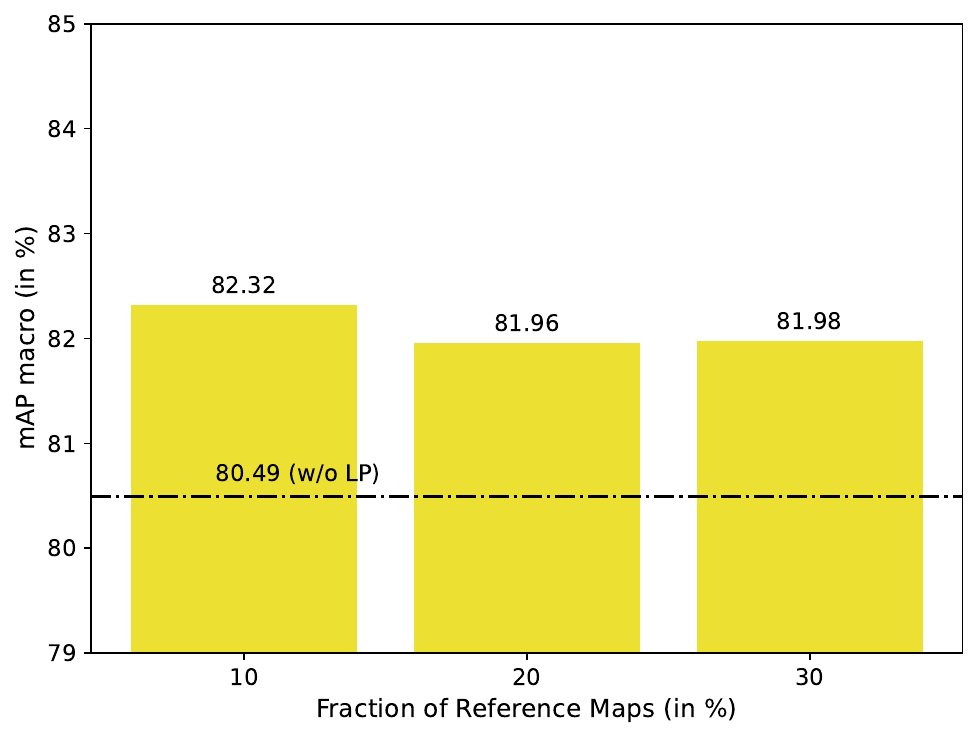}}
    \end{subfigure}
    \begin{subfigure}{\simnoisefrac\linewidth}
        \dynamiccaption{\caption{}\label{subfigure:sim_noise_class_swap_result}}{\includegraphics[width=\linewidth]{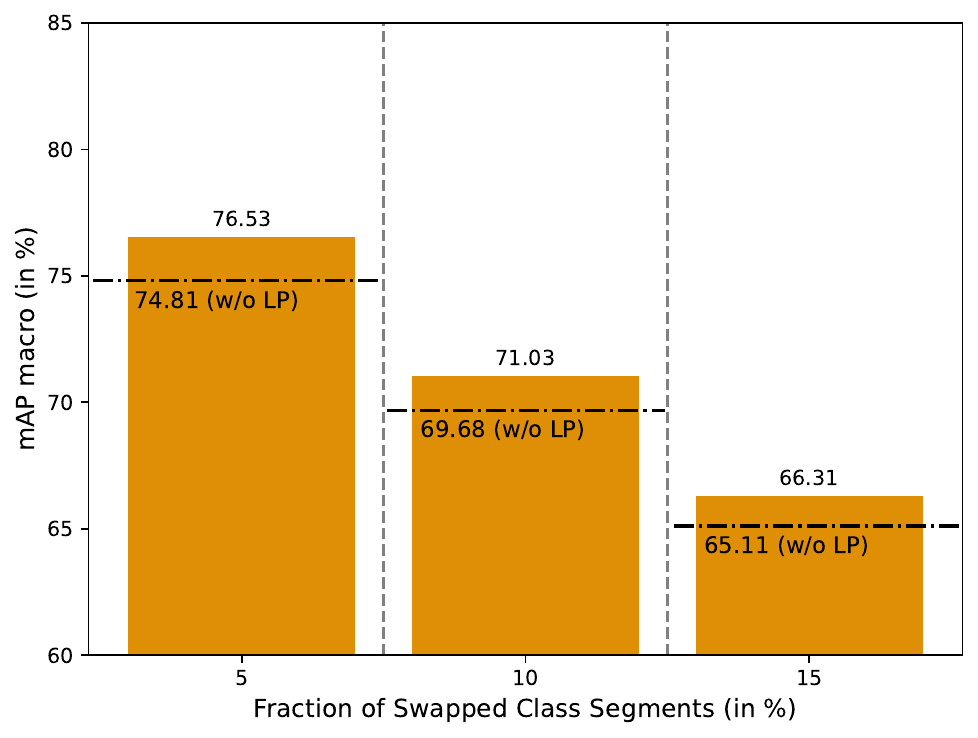}}
    \end{subfigure}
    \caption{Results of simulated noise in pixel-level reference maps for DeepGlobe-ML: \subref{subfigure:sim_noise_shift_result}~Mask Shift, \subref{subfigure:sim_noise_dilation_result}~Dilation/Erosion, \subref{subfigure:sim_noise_rectify_result}~Rectify Borders, \subref{subfigure:sim_noise_border_deform_result}~Border Deformation, \subref{subfigure:sim_noise_segment_swap_result}~Segment Swap, \subref{subfigure:sim_noise_class_swap_result}~Class Swap.}
\label{fig:sim_noise}
\end{figure*}

\Cref{fig:sim_noise} shows the results for different noise types in the reference maps of the DeepGlobe-ML dataset. All experiments with CutMix~w~$\text{LP}_\text{map}$ use the best box size range of \numrange{0.3}{0.7} with probability $p=0.5$, and are compared to CutMix w/o LP under the same setting (see Section \ref{subsec:clean_segmaps}). For Mask Shift, Dilation/Erosion, Rectify Borders, and Border Deformation, noise with varying intensity is applied to different fractions of the maps (\SI{25}{\percent}, \SI{50}{\percent}, \SI{100}{\percent}). For Mask Shift, our proposed \gls{LP} strategy consistently outperforms CutMix w/o LP (see \cref{subfigure:sim_noise_shift_result}). Shifts of 12 or 24 pixels (\SI{10}{\percent} and \SI{20}{\percent} of the height) barely affect performance, while a stronger shift of 36 pixels (\SI{30}{\percent}) only reduces accuracy if at least half of the reference maps contain noise. Even then, performance remains \SI{1.5}{\percent} above the baseline without any noise. In contrast, dilation or erosion reduces performance more clearly (see \cref{subfigure:sim_noise_dilation_result}), showing a linear decline with both the fraction of maps and $\alpha_\text{iteration}$. The lowest mAP macro (\SI{80.75}{\percent}) occurs at 36 iterations, where the results approach CutMix w/o LP. Under all other settings, our \gls{LP} strategy remains superior. Rectify Borders shows a similar but less pronounced linear trend (see \cref{subfigure:sim_noise_rectify_result}), with only minor performance drops, and our \gls{LP} strategy consistently outperforms CutMix w/o LP. Border Deformation has a negligible effect (see \cref{subfigure:sim_noise_border_deform_result}), with deviations of less than \SI{0.2}{\percent} in mAP macro, likely due to the high IoU preserved between noisy and clean reference maps. Segment Swap induces a clear linear relation between the fraction of moved segments and performance (see \cref{subfigure:sim_noise_segment_swap_result}). Even with \SI{30}{\percent} swaps, our \gls{LP} strategy still improves CutMix w/o LP by about \SI{1.5}{\percent}. For Class Swap, which introduces multi-label noise, CutMix w/o LP was retrained for each box size. Consistent with the clean setting, the \gls{LP} strategy yields a stable gain of about \SI{+1.5}{\percent} across all tested fractions (see \cref{subfigure:sim_noise_class_swap_result}). Overall, our \gls{LP} strategy remains more effective than CutMix without label updates under all noise conditions.

In summary, inaccurate borders due to border deformation or annotation errors in swapped class segments do not affect the effectiveness of our proposed \gls{LP} strategy. Some sensitivity is observed for shifts, low-resolution maps, dilation/erosion, and the replacement of class segments. In particular, strong dilation or erosion that affects more than \SI{50}{\percent} of the maps may reduce the performance to the level of CutMix w/o LP. In all other cases, performance degradations remain minor and it is still advantageous to apply our \gls{LP} strategy to CutMix. The robustness of the \gls{LP} strategy arises from the fact that rough positional estimates are sufficient to update the multi-label of an augmented image: if a class covers a larger area in one of the paired images, a single correctly annotated pixel can suffice for a correct update. The results indicate that the cases of wrong updates are less significant than the loss of information when our \gls{LP} strategy is not used.

\subsection{Label Propagation based on Noisy Reference Maps from Thematic Products} \label{subsec:thematic_products_segmaps}

To evaluate the effectiveness of our \gls{LP} strategy in a scenario in which class positional information is provided in the form of a thematic product, we adopt the same setup from Section \ref{subsec:clean_segmaps} for the BigEarthNet-S2 dataset. In line with the findings of our ablation study on simulated noisy reference maps, the use of our \gls{LP} strategy results in an improvement over CutMix without an effective label update (see \cref{table:results_ben19_thematic_product}). Among all box size ranges tested, CutMix~w~$\text{LP}_\text{map}$ consistently outperforms CutMix w/o LP. It is worth noting that the largest improvements are similar to those obtained for DeepGlobe-ML (see \ref{subsec:clean_segmaps}) (about \SI{2.0}{\percent} in mAP macro) even though the class positional information contains more noise due to the usage of reference maps derived from a thematic product. The noisy class positional information originating from a thematic product does not impact the effectiveness of our \gls{LP} strategy. In addition, unlike DeepGlobe-ML, the box size range does not seem to play an important role in the generalization performance. CutMix~w~$\text{LP}_\text{map}$ achieves the best results of about \SI{76.7}{\percent} in mAP macro for the box size ranges \numrange{0.1}{0.3}, \numrange{0.3}{0.5}, and \numrange{0.3}{0.7}. A clear trend in the reliance on smaller or larger boxes cannot be observed. 

\begin{table}[h!]
\caption{Results in mAP macro (\si{\percent}) for BigEarthnet-S2 with associated noisy reference maps derived from a thematic product.}
\label{table:results_ben19_thematic_product}
    \centering
    \begin{tabularx}{\linewidth}{X S S S S S}
     \toprule
     \multicolumn{6}{c}{\textbf{CutMix Augmentation}} \\
     \midrule
     & \multicolumn{5}{c}{Box Size Range} \\
     \cmidrule{2-6}
     Method & \numrange{0.1}{0.3} & \numrange{0.1}{0.5} & \numrange{0.1}{0.7} & \numrange{0.3}{0.5} & \numrange{0.3}{0.7} \\ 
     \midrule
     CutMix w/o LP & 75.42 & 74.90 & 74.69 & 74.52 & 75.39 \\ 
     {CutMix~w~$\text{LP}_\text{map}$} & \textbf{76.72} & \textbf{76.32} & \textbf{76.05} & \textbf{76.76} & \textbf{76.72} \\
     {CutMix~w~$\text{LP}_\text{xAI}$} & \textbf{77.21} & \textbf{77.22} & \textbf{77.21} & \textbf{78.26} & \textbf{78.43} \\
     \midrule
     \multicolumn{6}{c}{\textbf{No Augmentation}} \\
     \midrule
     Baseline & 71.46 & & & & \\
     \bottomrule
     \\
    \end{tabularx}
\end{table}

\subsection{Label Propagation based on Class Explanation Masks}\label{subsec:xai_segmaps}

\begin{figure}[h]
    \centering
    \begin{subfigure}{.48\linewidth}
        \dynamiccaption{\caption{}\label{subfigure:xai_example_original}}{\includegraphics[width=\linewidth]{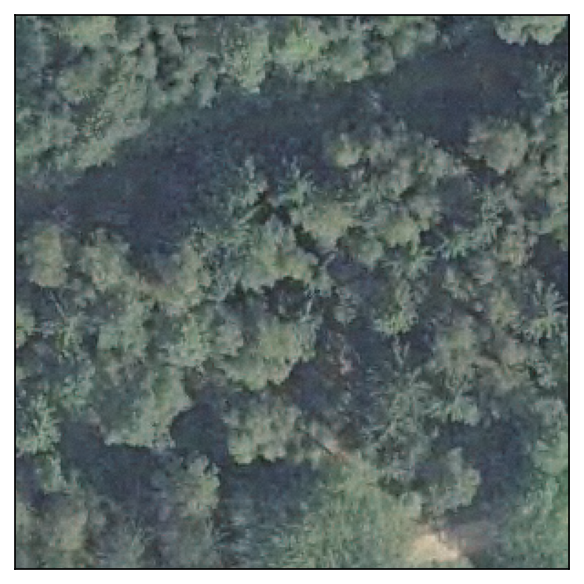}}
    \end{subfigure}
    \begin{subfigure}{.48\linewidth}
        \dynamiccaption{\caption{}\label{subfigure:xai_example_betula}}{\includegraphics[width=\linewidth]{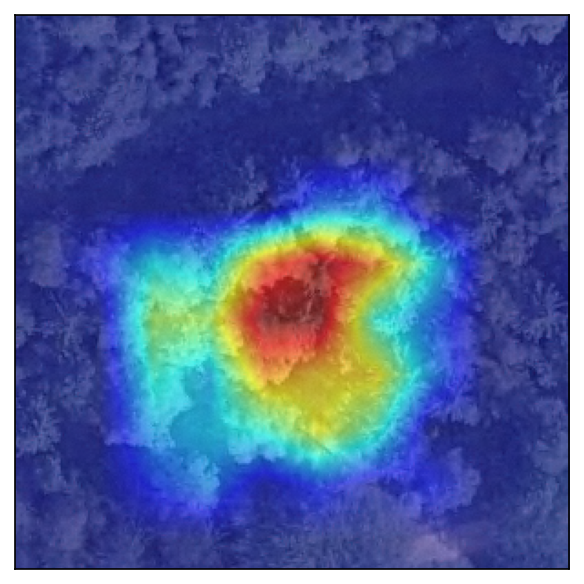}}
    \end{subfigure}
    \\\vspace{\subfigurerowdistance}
    \begin{subfigure}{.48\linewidth}
        \dynamiccaption{\caption{}\label{subfigure:xai_example_pinus}}{\includegraphics[width=\linewidth]{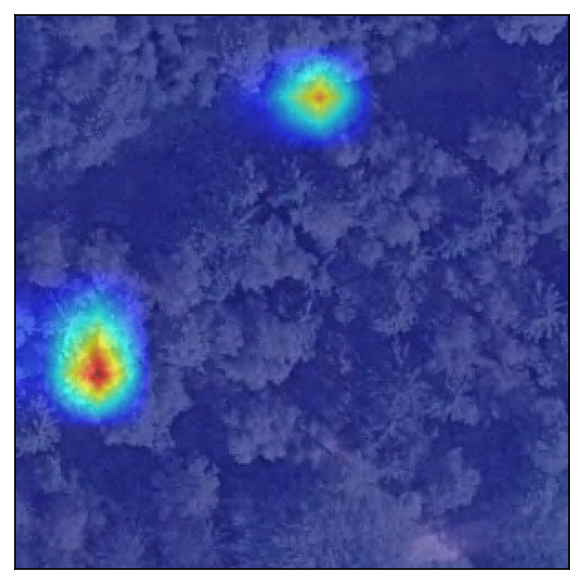}}
    \end{subfigure}
    \begin{subfigure}{.48\linewidth}
        \dynamiccaption{\caption{}\label{subfigure:xai_example_quercus}}{\includegraphics[width=\linewidth]{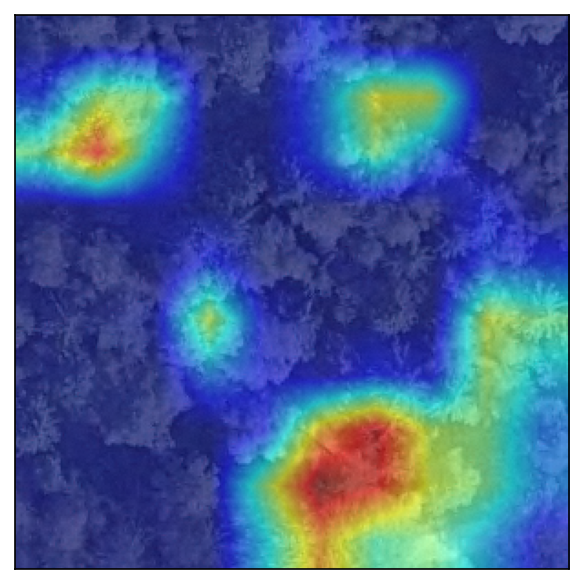}}
    \end{subfigure}
    \caption{Examples for CAMs generated by Grad-CAM for TreeSatAI: \subref{subfigure:xai_example_original}~original image, \subref{subfigure:xai_example_betula}~activation map for class Betula, \subref{subfigure:xai_example_pinus}~activation map for class Pinus, and \subref{subfigure:xai_example_quercus}~activation map for class Quercus.}
    \label{fig:xai_example}
\end{figure}

Finally, to show the generalization capability of our proposed \gls{LP} strategy to scenarios in which no associated pixel-level reference maps are available, we conduct experiments on TreeSatAI as well as on the DeepGlobe-ML and the BigEarthNet-S2 datasets, assuming the unavailability of their reference maps. In this experiment, we apply the \gls{LP} strategy to class explanation masks derived from an explanation method. To this end, we initially obtain the attribution of the importance of the pixels from \gls{CAMs} generated by the Grad-CAM method \cite{selvaraju_grad-cam_2017}. However, any other post hoc explanation method that can attribute multiple labels to the input pixels can be used. In this scenario, the workflow consists of two stages. First, a deep neural network is pre-trained on the task at hand until convergence without applying any augmentations. Then, a post hoc explanation method is applied to the pre-trained neural network to obtain the class explanation masks for each image. Subsequently, the actual training procedure that applies CutMix with our proposed \gls{LP} strategy starts with a new training procedure from scratch. As CAMs comprise heatmaps with continuous values, a threshold is applied to obtain binary masks that fulfill the requirements defined in Section \ref{mlc_cutmix} to be leveraged by our \gls{LP} strategy. For all experiments, we set the threshold $t_\text{cam}$ as 0.1, while we set the second threshold for minimum activating pixels $t_\text{map}$ as 10 to smooth uncertainty in border regions between activating and non-activating pixels. Nevertheless, it is worth noting that many combinations of parameters for both threshold parameters can lead to reasonable performance. We applied a grid search that indicated almost negligible differences. An example of \gls{CAMs} generated by GradCAM are shown in \cref{fig:xai_example}.

\begin{figure}
    \centering
        \includegraphics[width=0.8\linewidth]{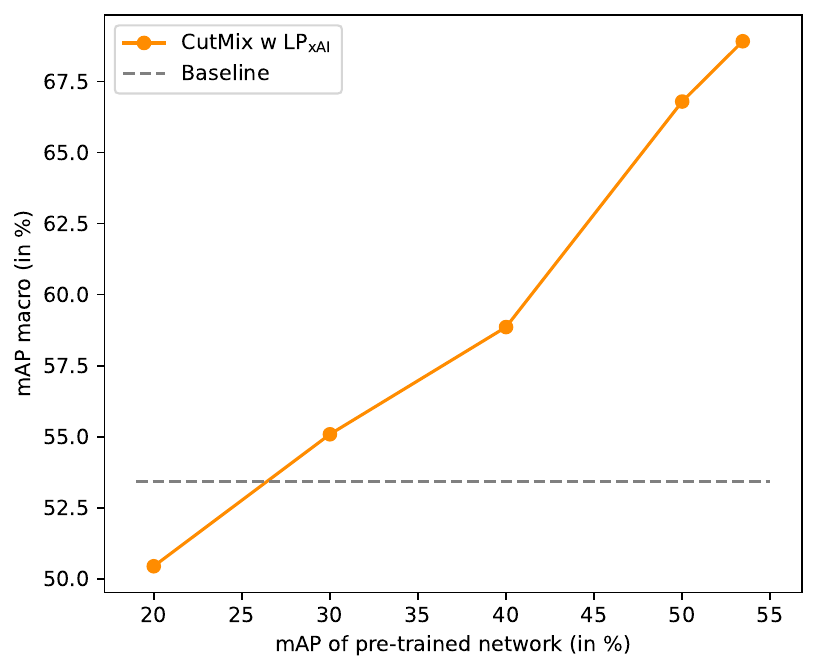}
    \caption{Ablation study on using pre-trained networks of varying levels of performance to generate class explanation masks for the CutMix~w~$\text{LP}_\text{xAI}$ training on TreeSatAI.}
    \label{fig:ablation_study}
\end{figure}

Regarding box sizes and application probability, we followed the same setup as in previous experiments. From \cref{table:results_treesatai_noisy_xai}, it can be observed that the results of TreeSatAI exhibit similar trends as with reliable reference maps. Even though TreeSatAI is the most challenging dataset and the class positional information is derived from noisy CAMs, the observations from reliable reference maps can be generalized to scenarios in which no associated reference maps are available. In line with the results of \ref{subsec:noisy_segmaps}, not fully reliable class positional information provided by explanation methods does not reduce the effectiveness of our proposed \gls{LP} strategy. For all box size ranges tested, CutMix~w~$\text{LP}_\text{xAI}$ outperforms its counterpart without \gls{LP}. Similarly, as observed for reliable reference maps, the more larger boxes are used, the better CutMix~w~$\text{LP}_\text{xAI}$ becomes, leading to the best mAP macro of \SI{68.91}{\percent} for a box size range of \numrange{0.3}{0.7}. The effectiveness of using \gls{LP} strategy based on class explanation masks is also observable for DeepGlobe-ML and BigEarthNet-S2 (see CutMix~w~$\text{LP}_\text{xAI}$ in \cref{table:results_deepglobe_clean} and \cref{table:results_ben19_thematic_product}), respectively. It is noteworthy that class explanation masks yield results that are almost on-par with reliable reference maps, and even better results than not fully reliable reference maps. Furthermore, an ablation study comparing class explanation masks derived from pre-trained networks with varying levels of performance shows that our \gls{LP} strategy based on class explanation masks is already effective when the class explanation masks are obtained from pre-trained models that are less performant. \cref{fig:ablation_study} shows the results of the \gls{LP} strategy based on class explanation masks obtained from intermediate model states saved during pre-training (stage one) on TreeSatAI with a fixed box size range of \numrange{0.3}{0.7}. It can be seen that already at a pre-training mAP macro of \SI{30}{\percent} the produced explanation masks provide enough reliable class positional information such that CutMix~w~$\text{LP}_\text{xAI}$ outperforms the baseline without augmentation. Consequently, our \gls{LP} strategy also demonstrates effectiveness for scenarios in which explanation methods are used to obtain class positional information to allow the effective use of CutMix in \gls{MLC}.

\begin{table}[h!]
\caption{Results in mAP macro (\si{\percent}) for TreesSatAI with associated class explanation masks obtained by the thresholded \gls{CAMs}.}
\label{table:results_treesatai_noisy_xai}
    \centering
     \begin{tabularx}{\linewidth}{X S S S S S S}
     \toprule
     \multicolumn{6}{c}{\textbf{CutMix Augmentation}} \\
     \midrule
     & \multicolumn{5}{c}{Box Size Range} \\
     \cmidrule{2-6}
     Method & \numrange{0.1}{0.3} & \numrange{0.1}{0.5} & \numrange{0.1}{0.7} & \numrange{0.3}{0.5} & \numrange{0.3}{0.7} \\ 
     \midrule
     CutMix w/o LP & 57.42 & 58.78 & 60.19 & 63.82 & 64.67 \\ 
     {CutMix~w~$\text{LP}_\text{xAI}$} & \textbf{62.04} & \textbf{63.51} & \textbf{64.48} & \textbf{68.77} & \textbf{68.91} \\
     \midrule
     \multicolumn{6}{c}{\textbf{No Augmentation}} \\
     \midrule
     Baseline & 53.44 & & & & \\
     \bottomrule
     \\
     
    \end{tabularx}
\end{table}

\subsection{Comparison with other Data Augmentation Techniques} \label{subsec:other_das}

To contextualize the effectiveness of our proposed \gls{LP} strategy within the broader landscape of other data augmentation techniques, we compare its performance against standard augmentation methods, including CutOut, Random Resize Crop, Random Rotate, and Mixup. The results of these comparisons on the DeepGlobe-ML dataset are presented in \cref{table:results_data_aug_comparison}. The baseline model, trained without any augmentation, achieves an AP macro of \SI{78.40}{\percent}. Among the conventional augmentation techniques, Random Resize Crop provides the highest improvement, yielding an AP macro of \SI{81.41}{\percent}. When we examine the impact of CutMix, we observe a notable performance boost. Without our \gls{LP} strategy, CutMix achieves an AP macro of \SI{80.49}{\percent}, already surpassing CutOut, Mixup, and Random Rotate. However, when our \gls{LP} strategy is incorporated, the improvements become even more pronounced. CutMix~w~$\text{LP}_\text{map}$ achieves the best overall performance, reaching \SI{82.30}{\percent} in AP macro, outperforming all other augmentation strategies. Similarly, CutMix~w~$\text{LP}_\text{xAI}$ achieves an AP macro of \SI{82.01}{\percent}, demonstrating the effectiveness of leveraging class positional information derived from explanation methods. These results confirm that the proposed \gls{LP} strategy not only enhances the effectiveness of CutMix but also outperforms commonly used geometric augmentation techniques. The significant improvements in AP macro underscore the importance of accurate multi-label propagation in \gls{MLC}, particularly when class positional information is available. Moreover, the gains observed with CutMix~w~$\text{LP}_\text{xAI}$ highlight the feasibility of employing explanation-driven class positional information in cases where explicit reference maps are unavailable.

\begin{table}[h!]
\centering
\renewcommand{\arraystretch}{1.2} 
\setlength{\tabcolsep}{12pt} 
\caption{Comparison of CutMix using our proposed \gls{LP} strategy with other geometric data augmentation techniques for DeepGlobe-ML.}
\label{table:results_data_aug_comparison}
\begin{tabularx}{0.9\linewidth}{>{\arraybackslash}X S[table-format=2.2] S[table-format=2.2]}
\toprule
    \textbf{Method} & \textbf{mAP macro} & \textbf{mAP micro} \\
    \midrule
    Baseline                  & 78.40 & 82.90 \\
    CutOut                    & 79.69 & 84.93 \\ 
    RandomResizeCrop          & 81.41 & 86.72 \\
    RandomRotate              & 78.55 & 83.75 \\
    Mixup                     & 77.45 & 83.75 \\
    CutMix w/o LP             & 80.49 & 83.65 \\ 
    \midrule
    CutMix~w~$\text{LP}_\text{map}$ & \textbf{82.30} & \textbf{87.25}\\
    CutMix~w~$\text{LP}_\text{xAI}$ & 82.01 & 86.88 \\
    \bottomrule
\end{tabularx}
\end{table}
\section{Conclusion}\label{conclusion}

In this article, we have proposed a label propagation (LP) strategy that enables the effective application of CutMix for multi-label image classification (MLC) in remote sensing (RS). Directly applying CutMix to multi-label training images can cause erasure or addition of class labels in the paired image and, thus, introduce multi-label noise. The proposed \gls{LP} strategy mitigates this issue by preserving correct class information within the multi-label vector of the augmented image. It exploits pixel-level class positional information to ensure accurate label updates for erased and copied areas. This information can originate either from associated reference maps or from class explanation masks derived by explanation methods. Similar to the pairing operation of CutMix for two images, our \gls{LP} strategy combines the corresponding pixel-level class information from both images to derive the updated multi-label vector. Experimental results demonstrate the effectiveness of the \gls{LP} strategy under both reliable reference maps and simulated noisy reference maps with less accurate class positional information. In line with these findings, the \gls{LP} strategy also performs well in practical scenarios using thematic products or class explanation masks as the source for class positional information. Its robustness to imperfect class positional information arises from the fact that, in MLC tasks, approximate localization is often sufficient for maintaining correct multi-label assignments. Consequently, integrating the proposed LP strategy allows CutMix to be applied in RS MLC without introducing label noise.

Although the LP strategy is in principle applicable to other augmentations that omit parts of the image, such as CutOut or random cropping, we did not observe significant performance gains in these settings. This may be explained by the fact that such techniques primarily introduce additive label noise, which is less detrimental in MLC tasks than subtractive noise, as shown in previous work \cite{burgert_effects_2022}. Since CutMix uniquely introduces subtractive noise by inserting content from other images, it is particularly well suited for correction via our LP strategy.

As a final remark, we would like to note that the proposed strategy is promising for possible operational applications in which associated class positional information might not be fully reliable, due to both its general properties and its simplicity in implementation. Our experiments demonstrate that while the LP strategy remains effective under imperfect or noisy class positional information, its performance can benefit from more reliable inputs, such as well-aligned and temporally consistent reference maps. We therefore recommend, whenever possible, validating or cross-checking the quality of thematic products used to generate pixel-level information. In cases where such products are unavailable or of low quality, explanation masks offer a practical alternative, allowing the strategy to operate effectively without explicit reference maps. During pretraining, we intentionally omitted data augmentation to obtain a controlled baseline for generating class explanation masks, which allowed us to attribute performance gains more clearly to the proposed label propagation. In operational scenarios, we recommend advanced augmentation and regularization during pretraining to obtain the strongest possible backbone. The modular pipeline also permits future extensions toward an online version that leverages dynamically computed explanation masks.

\section{Acknowledgement}
This work is supported by the European Research Council (ERC) through the ERC-2017-STG BigEarth Project under Grant 759764.

\ifCLASSOPTIONcaptionsoff
  \newpage
\fi

\bibliography{references}
\bibliographystyle{ieeetr}
\vfill

\begin{IEEEbiography}[{\vspace{-17pt}\includegraphics[width=1in,height=1.25in,clip,keepaspectratio]{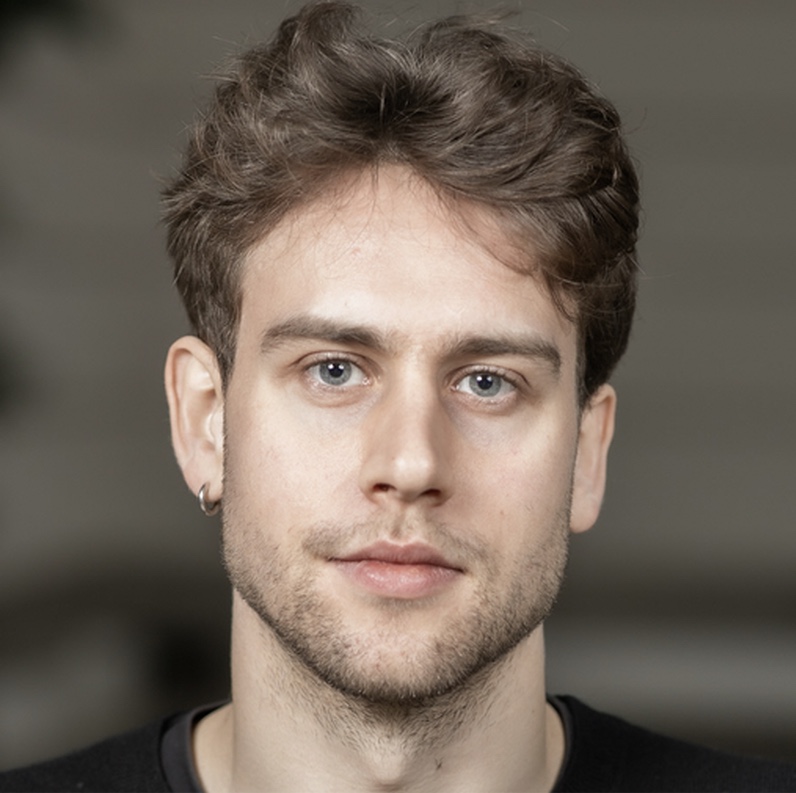}}]{Tom Burgert} received his M.Sc. degree in computer science from Technische Universit\"at (TU) Berlin in 2022. Currently, he is a first-year Ph.D. student in machine learning at the Remote Sensing and Image Analysis (RSiM) Group at the TU Berlin. He joined the RSiM Group as a student research assistant during his postgraduate study. His research interests evolve around learning theory and explaining deep neural networks in the intersection of computer vision and remote sensing. \end{IEEEbiography}

\begin{IEEEbiography}[{\vspace{-17pt}\includegraphics[width=1in,height=1.25in,clip,keepaspectratio]{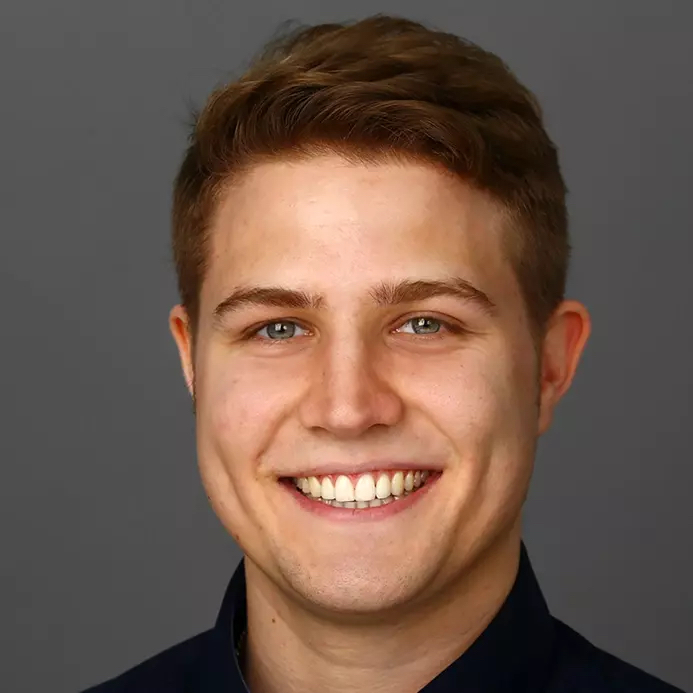}}]{Kai Norman Clasen} received the B.Sc. and M.Sc. degrees in computer engineering from Technische Universität Berlin, Berlin, Germany, in 2018 and 2020, respectively, where he is currently pursuing the Ph.D. degree with the Remote Sensing Image Analysis (RSiM) Group, Faculty of Electrical Engineering and Computer Science, and the Big Data Analytics for Earth Observation Research Group, Berlin Institute for the Foundations of Learning and Data (BIFOLD), Berlin. His research interests revolve around the intersection of remote sensing and deep learning, and he has a particular interest in reproducibility and open science. \end{IEEEbiography}

\begin{IEEEbiography}[{\vspace{-17pt}\includegraphics[width=1in,height=1.25in,clip,keepaspectratio]{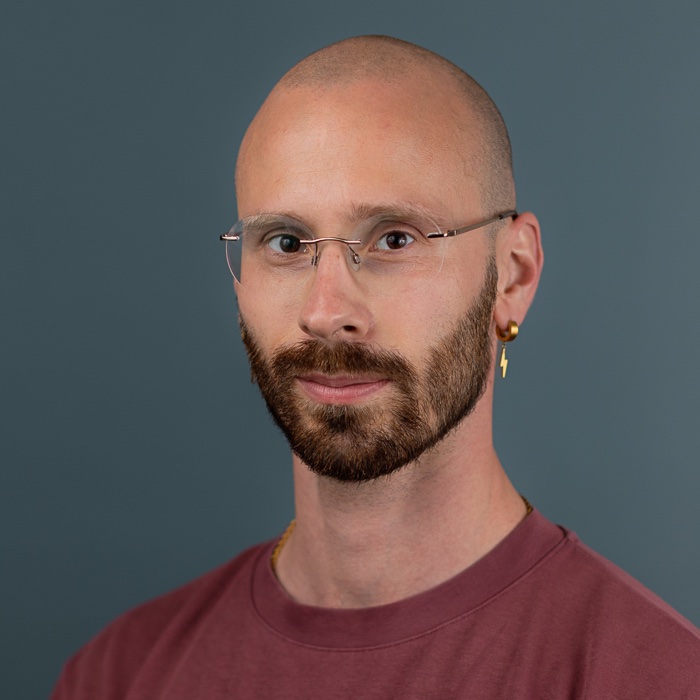}}]{Jonas Klotz} received the M.Sc. degree in computer science from Technische Universität (TU) Berlin, Berlin, Germany, in 2024. He is currently pursuing a Ph.D. in machine learning at the Berlin Institute for the Foundations of Learning and Data (BIFOLD), with the Remote Sensing and Image Analysis (RSiM) Group. His research interests center on the intersection of explainable AI and remote sensing.
\end{IEEEbiography}

\begin{IEEEbiography}[{\vspace{-17pt}\includegraphics[width=1in,height=1.25in,clip,keepaspectratio]{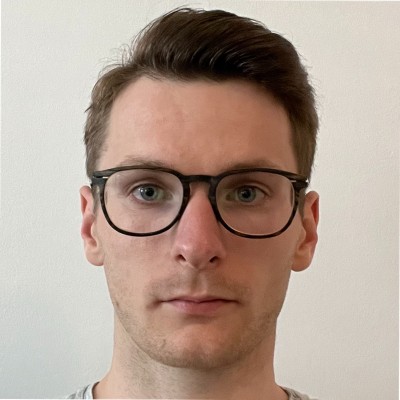}}]{Tim Siebert} received the M.Sc. degree in computer science from Technische Universität (TU) Berlin, Berlin, Germany, in 2024. He is currently pursuing a Ph.D. degree at Humboldt-Universität zu Berlin, where his research focuses on mathematical foundations of machine learning, in particular efficient computation of partial differential equation operators and automatic differentiation. His broader interests include scientific machine learning, optimization, and compiler-assisted differentiation techniques.
\end{IEEEbiography}

\begin{IEEEbiography}[{\vspace{-17pt}\includegraphics[width=1in,height=1.25in,clip,keepaspectratio]{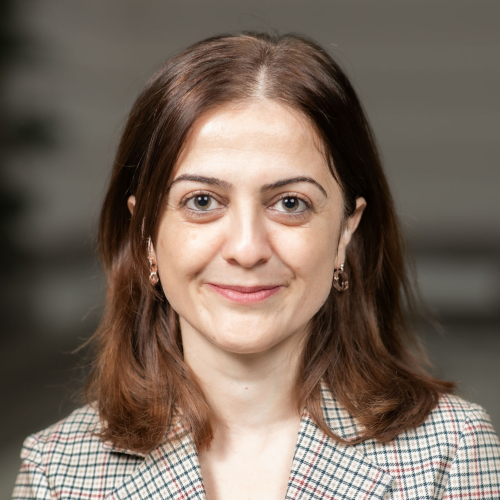}}]{Begüm Demir} (demir@tu-berlin.de) (S'06-M'11-SM'16) received the B.Sc., M.Sc., and Ph.D. degrees in electronic and telecommunication engineering from Kocaeli University, Kocaeli, Turkey, in 2005, 2007, and 2010, respectively.
She is currently a Full Professor and the founder head of the Remote Sensing Image Analysis (RSiM) group at the Faculty of Electrical Engineering and Computer Science, TU Berlin and the head of the Big Data Analytics for Earth Observation research group at the Berlin Institute for the Foundations of Learning and Data (BIFOLD). Her research activities lie at the intersection of machine learning, remote sensing and signal processing. Specifically, she performs research in the field of processing and analysis of large-scale Earth observation data acquired by airborne and satellite-borne systems. She was awarded by the prestigious ‘2018 Early Career Award’ by the IEEE Geoscience and Remote Sensing Society for her research contributions in machine learning for information retrieval in remote sensing. In 2018, she received a Starting Grant from the European Research Council (ERC) for her project “BigEarth: Accurate and Scalable Processing of Big Data in Earth Observation”. She is an IEEE Senior Member and Fellow of European Lab for Learning and Intelligent Systems (ELLIS).
Prof. Demir is a Scientific Committee member of several international conferences and workshops. She is a referee for several journals such as the PROCEEDINGS OF THE IEEE, the IEEE TRANSACTIONS ON GEOSCIENCE AND REMOTE SENSING, the IEEE GEOSCIENCE AND REMOTE SENSING LETTERS, the IEEE TRANSACTIONS ON IMAGE PROCESSING, Pattern Recognition, the IEEE TRANSACTIONS ON CIRCUITS AND SYSTEMS FOR VIDEO TECHNOLOGY, the IEEE JOURNAL OF SELECTED TOPICS IN SIGNAL PROCESSING, the International Journal of Remote Sensing), and several international conferences. Currently she is an Associate Editor for the IEEE GEOSCIENCE AND REMOTE SENSING MAGAZINE.
\end{IEEEbiography}

\end{document}